%% file: main.tex
\documentclass{article}

\usepackage[preprint]{neurips_2023}
\usepackage{times}
\usepackage{epsfig}
\usepackage{graphicx}
\usepackage{wrapfig}
\usepackage{amsmath}
\usepackage{amssymb}
\usepackage{multicol}
\usepackage{xcolor} % colors
\usepackage{tablefootnote}
\usepackage{diagbox}
\usepackage{subfigure} % For subfigures
\usepackage{multirow}

\newcommand{\highlight}[1]{\colorbox{blue!10}{#1}}
\definecolor{mygray}{gray}{0.4}

\newcommand{\g}[2]{#1\textsubscript{\textcolor{mygray}{$\pm$#2}}}

\begin{document}

%%%%%%%%% TITLE
\title{Cycle Consistency Driven Object Discovery}

\author{
Aniket Didolkar \textsuperscript{1}, 
Anirudh Goyal \textsuperscript{2}, 
\textbf{Yoshua Bengio \textsuperscript{1}}}

\maketitle
%\ificcvfinal\thispagestyle{empty}\fi
\let\footnote\relax\footnotetext{ \textsuperscript{1} Mila, University of Montreal, \textsuperscript{2} Google DeepMind \\
Corresponding authors:  \texttt{adidolkar123@gmail.com} 
}
%%%%%%%%% ABSTRACT
\begin{abstract}

Developing deep learning models that effectively learn object-centric representations, akin to human cognition, remains a challenging task. Existing approaches facilitate object discovery by representing objects as fixed-size vectors, called ``slots'' or ``object files''. While these approaches have shown promise in certain scenarios, they still exhibit certain limitations. First, they rely on architectural priors which can be unreliable and usually require meticulous engineering to identify the correct objects. Second, there has been a notable gap in investigating the practical utility of these representations in downstream tasks. To address the first limitation, we introduce a method that explicitly optimizes the constraint that each object in a scene should be associated with a distinct slot. We formalize this constraint by introducing  consistency objectives which are cyclic in nature. By integrating these consistency objectives into various existing slot-based object-centric methods, we showcase substantial improvements in object-discovery performance. These enhancements consistently hold true across both synthetic and real-world scenes, underscoring the effectiveness and adaptability of the proposed approach. To tackle the second limitation, we apply the learned object-centric representations from the proposed method to two downstream reinforcement learning tasks, demonstrating considerable performance enhancements compared to conventional slot-based and monolithic representation learning methods. Our results suggest that the proposed approach not only improves object discovery, but also provides richer features for downstream tasks.

\end{abstract}

%%%%%%%%% BODY TEXT
\section{Introduction}

Having object-based reasoning capabilities is crucial for solving many real-world problems. The world around us is complex, diverse, and filled with distinct objects. Human beings naturally possess the ability to parse and reason about these objects in their environment. Frequently, changing or manipulating certain aspects of the world requires interacting with a single object or a combination of objects. For instance, driving a car necessitates maneuvering a single object (the car) while avoiding collisions with other objects or entities, such as other cars, trees, and obstacles. Developing object-based reasoning capabilities is therefore crucial for improving the ability of deep learning models to understand and solve problems in the real world.

The unsupervised discovery of objects from a scene poses a challenging problem, as determining what an object refers to can be difficult to parse without additional context. Many existing approaches \citep{greff2017nem, greff2019iodine, burgess2019monet, goyal2019recurrent, slot-attention, goyal2020object, goyal2021neural, ke2021systematic, goyal2021coordination, singh2022illiterate} learn a set of slots to represent objects, where each slot is a fixed-size vector. Most of these approaches use a reconstruction loss combined with specific architectural biases that rely on visual cues to segment objects into slots. Some work has utilized other auxiliary cues for supervision, such as optical flow \citep{kipf2021conditional} and depth prediction \citep{elsayed2022savi}. However, architectural priors may not always be reliable and hence may not scale to real-world data. Relying on auxiliary information like optical flow and motion is not feasible since many datasets and scenes lack this information. To address these limitations, we augment existing slot-based methods with two auxiliary objectives called 'cycle consistency' objectives  %Instead, the proposed approach relies on explicitly enforcing coherence and consistency between the representations of the visual features (obtained from a convolutional or transformer-based encoder) and the learned slots. The proposed approach is simple and flexible, and it can be integrated into any slot-based object-discovery method.

\begin{figure}
    \centering
    \includegraphics[width = \linewidth]{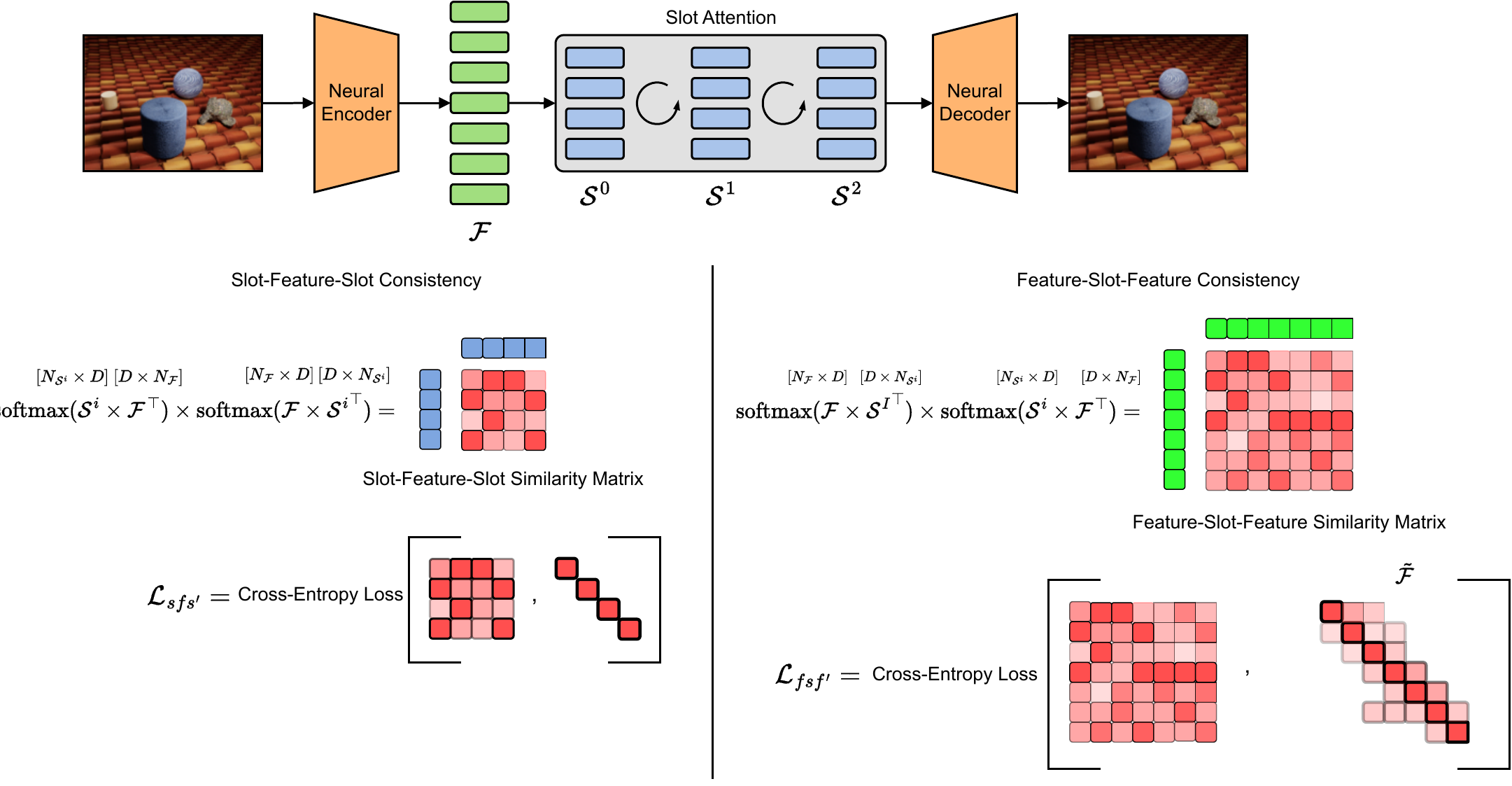}
    \caption{\textbf{Cycle Consistency Objectives}. Here, we present the general architecture of the model. The proposed approach augments two additional losses to existing object-centric methods (specifically, slot attention in this case). These losses directly encourage object discovery within the latent space. We refer to these losses as the \textsc{Slot-Feature-Slot} Consistency loss and the \textsc{Feature-Slot-Feature} consistency loss. The \textsc{Slot-Feature-Slot} consistency loss is calculated as the cross-entropy between the slot-feature-slot similarity matrix and the identity matrix. The \textsc{Feature-Slot-Feature} consistency loss is computed as the cross-entropy between the feature-slot-feature similarity matrix and a custom matrix $\tilde{\mathcal{F}}$, the details of which are provided in Section \ref{sec:method}.}
    \label{fig:method_demo}
\end{figure}

The proposed cycle consistency objectives operate directly on the latent representations of the slots and visual features (obtained from the neural encoder, as shown in Figure \ref{fig:method_demo}). These objectives augment the architectural priors used in slot attention-style models with an additional layer of reliability by explicitly enforcing coherence and consistency between the representations of the visual features and the learned slots. To apply the objectives, we consider the visual features and slots as nodes in a directed graph. The problem of object discovery can then be formulated as the task of adding the correct edges into the graph such that: (1) the outgoing edges from a set of features belonging to the same object should lead to the same slot, and (2) the outgoing edges from each slot should connect to a distinct subset of features. Both these constraints are formulated into two cycle consistency objectives called the \textsc{Slot-Feature-Slot} consistency loss and the \textsc{Feature-Slot-Feature} consistency loss. Further details regarding these objectives are elaborated in Section \ref{sec:method}.

The proposed objectives draw motivation from similar objectives presented in \citep{wang2023objectcentric}. We propose modifications that demonstrate the effectiveness of these objectives across multiple downstream RL tasks and various object-discovery settings, different from those studied in \citep{wang2023objectcentric}. These objectives are simple and can seamlessly integrate into any existing slot-based object-discovery method. Our findings indicate that augmenting slot-based methods with these proposed objectives enhances object-discovery performance and exhibits stronger generalization to unseen scenarios. The learned representations also demonstrate significant transferability to various downstream RL tasks compared to several baselines.

\section{Proposed Method} \label{sec:method}

In this section, we present the details of the proposed cycle consistency objectives and the underlying intuition behind them. The proposed method is designed to operate on a set of $N$ slots, denoted as $\mathcal{S} = {s_0, s_1, \ldots, s_N}$, and a set of $M$ features, denoted as $\mathcal{F} = {f_0, f_1, \ldots, f_M}$, which can be obtained using any suitable backbone, such as a convolutional encoder. Similarly, the slots $\mathcal{S}$ can be obtained using any suitable object discovery or slot extractor method, such as slot attention, as illustrated in Figure \ref{fig:method_demo}:

\begin{equation}
\mathcal{S} = \text{slot\_extractor}(\mathcal{F})
\end{equation}

\paragraph{Preliminary Setup}
We denote the directed graph between the nodes and the features as $\mathcal{G}$. To determine the correct edges to add to this graph, we initially score all possible edges. Subsequently, we optimize the scores to satisfy the following two conditions:
\begin{itemize}
\item A set of features belonging to the same object must have outgoing edges to the same slot.
\item Outgoing edges from each slot should connect to a distinct set of features.
\end{itemize}

The score for an edge between a slot and a feature is computed by taking the dot product of their respective features. For instance, the score for an outgoing edge from feature $f_i$ to slot $s_j$ is calculated as follows:
\begin{equation}
    \phi(f_i, s_j) = \phi(f_i \rightarrow s_j) = \frac{f_i \cdot s_j}{\tau_1}
\end{equation}

Here, $\tau_1$ is the temperature which is generally set to 0.1 in our experiments. We convert these scores into probabilities by normalizing across all possible target nodes.

\begin{equation}
p(f_i \rightarrow s_j) = \frac{\exp(\phi(f_i \rightarrow s_j))}{\sum_{k = 0}^{k = N - 1}\exp(\phi(f_i \rightarrow s_k))}
\end{equation}

By computing these probabilities for all possible feature-slot pairs, we obtain a feature-slot similarity matrix $A_{f\rightarrow s} \in \mathbb{R}^{M \times N}$. Similarly, we compute a slot-feature similarity matrix $A_{s \rightarrow f} \in \mathbb{R}^{N \times M}$. Note that the scores in $A_{s \rightarrow f}$ are normalized across all features.

We aim for both similarity matrices to adhere to the aforementioned conditions. For instance, each row in $A_{s \rightarrow f}$, representing a specific slot $s_i$, should assign the highest probability to the features corresponding to that slot. Similarly, each row in $A_{f \rightarrow s}$ should assign the highest probability to the slots that correspond to the feature.

However, we cannot directly optimize these similarity matrices to meet these conditions because we lack the ground truth slot-to-feature assignments. Instead, we consider cyclic paths consisting of two edges: (1) \textsc{Slot-Feature-Slot} paths—paths with an edge from a slot to a feature and another edge from a feature to a slot; (2) \textsc{Feature-Slot-Feature} paths—paths with an edge from a feature to a slot and another edge from a slot to a feature.

\paragraph{\textsc{Slot-Feature-Slot} Consistency Loss} We desire that for an outgoing edge from slot $s_i$ to feature $f_k$, the outgoing edge from feature $f_k$ should cycle back to slot $s_i$. To gain intuition about this, consider a scenario where perfect object factorization has been achieved, where each object is represented by a distinct slot. In such a scenario, for an edge going from slot $s_i$ to feature $f_k$, the outgoing edge from $f_k$ will always return to $s_i$ because in the case of perfect factorization, each feature belonging to a particular object will be represented by only one slot—the slot that represents that particular object.

To impose this constraint, we first calculate the scores of all possible \textsc{Slot-Feature-Slot} paths, i.e., paths with edges from a slot $s_i$ to a feature $f_k$ and another edge from $f_k$ to slot $s_j$, as follows: $\phi(s_i \rightarrow f_k \rightarrow s_j) = p(s_i \rightarrow f_k) \cdot p(f_k \rightarrow s_j)$. We convert this into probabilities by normalizing over all possible paths beginning from slot $s_i$ and ending in slot $s_j$ as follows:
\begin{align}
    A_{s \rightarrow f \rightarrow s}[i, j] &= \frac{\exp(p(s_i \rightarrow f_k)\cdot p(f_k \rightarrow s_j))}{\sum_{k' = 0}^{M - 1} \sum_{j' = 0}^{N-1} \exp(p(s_i \rightarrow f_{k'})\cdot p(f_{k'} \rightarrow s_{j'}))} \\
   A_{s \rightarrow f \rightarrow s} &= \text{softmax}(A_{s\rightarrow f} A_{f \rightarrow s}, axis = 1)
\end{align}

We refer to $A_{s \rightarrow f \rightarrow s} \in \mathbb{R}^{N \times N}$ as the slot-feature-slot similarity matrix as shown in the Figure \ref{fig:method_demo}. Since we want every path originating at slot $s_i$ to return back to slot $s_i$, we want the probabilities along the diagonal of $A_{s \rightarrow f \rightarrow s}$ to be the highest hence we frame the \textsc{Slot-Feature-Slot} consistency loss as - 

\begin{equation}
    \mathcal{L}_{sfs'} = -\sum_{i = 0}^{N-1}\log(A_{s \rightarrow f \rightarrow s}[i, i]) 
\end{equation}

\paragraph{\textsc{Feature-Slot-Feature} Consistency Loss} For a path originating at a feature $f_i$, belonging to an object $o$, going into a slot $s_k$. The outgoing edge from slot $s_k$ must go into any of the features $f_j$ that represent $o$. Note that in this case the outgoing path from slot $s_k$ may not go back to the originating feature $f_i$ since multiple features belonging to one object map to the same slot.

To achieve this constraint, we first calculate the feature-slot-feature similarity matrix which is shown in Figure \ref{fig:method_demo} as follows - 

\begin{equation}
    A_{f \rightarrow s \rightarrow f} = \text{softmax}(A_{f \rightarrow s}  A_{s \rightarrow f}, axis = 1)
\end{equation}

Note that $A_{f \rightarrow s \rightarrow f} \in \mathbb{R}^{M \times M}$. In this case, we cannot optimize for the diagonals of $A_{f \rightarrow s \rightarrow f}$ to have the highest probabilities since \textsc{Slot-Feature-Slot} paths can have different source and target nodes. Therefore, we optimize it by computing the cross-entropy between $A_{f \rightarrow s \rightarrow f}$ and a custom matrix $\tilde{\mathcal{F}}$ - 

\begin{equation}
    \mathcal{L}_{fsf'} = -\sum_{i = 0}^{M-1}\tilde{\mathcal{F}}[i, i]\log(A_{f \rightarrow s \rightarrow f}[i, i]) 
\end{equation}

Note that the above loss is \textbf{only} computed for the diagonal elements of $A_{f \rightarrow s \rightarrow f}$. 

$\tilde{\mathcal{F}}$ is calculated as a function of the features $\mathcal{F}$ output by the encoder as indicated in Figure \ref{fig:method_demo}. First we calculate the pairwise feature similarity values using $\mathcal{F}$ and sparsify the feature similarity matrix based on a threshold $T$. Consider two features from $\mathcal{F}$ - $f_i$ and $f_j$. The similarity score between these features is calculated as $\delta_{i, j} = \frac{f_i \cdot f_j}{\tau_2}$. The threshold value $T$ is computed as  $T = c \cdot (\max(\mathcal{F}) - \min(\mathcal{F})) + \min(\mathcal{F})$, where $c$ is a hyperparameter. In all our experiments, we set $c$ to 0.8 unless specified otherwise. Once we obtain the sparse feature similarity matrix (denoted as $\tilde{F}$ which is a $M \times M$ matrix), we normalize it across rows to convert the similarity scores into probabilities - 

\begin{equation}
    \tilde{\mathcal{F}} = \text{softmax}(\tilde{F}, axis = 1)
\end{equation}

\paragraph{Training Details}
The proposed method can be applied on top of any slot-based object discovery method. To integrate this approach, we include the cycle consistency objectives into the original loss of the method. For instance, slot attention \citep{slot-attention} utilizes a pixel-based reconstruction loss. Upon adding the proposed objectives, the final loss becomes:

\begin{equation}
\mathcal{L}{final} = \mathcal{L}{recon} + \lambda_{sfs'} \mathcal{L}{sfs'} + \lambda{fsf'} \mathcal{L}_{fsf'}
\end{equation}

To set the hyperparameters for our approach, we select $\lambda_{sfs'}$ and $\lambda_{fsf'}$ as 0.1 and 0.01, respectively, unless specified otherwise. Additionally, we employ an Exponential Moving Average (EMA) visual encoder. The EMA encoder calculates $\tilde{\mathcal{F}}$. This practice aligns with several self-supervised learning studies \citep{moco, simsiam, byol}, ensuring that the features used in computing $\tilde{\mathcal{F}}$ remain stable, avoiding frequent changes caused by gradient updates in the visual encoder. Moreover, we detach the calculation of $\tilde{\mathcal{F}}$ from the gradient computation.

In methods involving multiple iterations of object discovery, such as slot attention, we apply the cycle consistency objectives to the slots obtained from each iteration of the method, unless explicitly stated otherwise. We set $\tau_1$ to 0.1 and $\tau_2$ to 0.01 unless specified differently.

%{\color{red} \paragraph{Connections to k-means clustering} \citep{slot-attention} mention that slot attention can be seen as a form soft k-means clustering with dot product as the similarity function and a learned update rule. Taking this view, the proposed cycle consistency objectives can be seen as a further enforcing function on the clustering that acts by maximizing the similarity between features that belong to a cluster and minimizing the similarity of features that belong to seperate clusters.  }

\begin{figure}
    \centering
    \includegraphics[width = \linewidth]{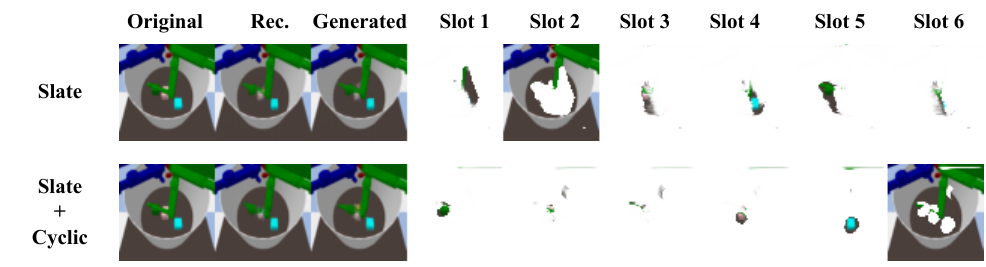}
    \caption{\textbf{Causal World Slot Masks} In this figure, we show the learned slot masks for the models i.e. with and without the proposed objecgives. We can see that proposed approach captures obejcts more clearly than the baseline Slate model.}
    \label{fig:cw_vis}
\end{figure}

\section{Related Work}
\paragraph{Unsupervised Object Discovery} Our work addresses the problem of object discovery in visual scenes. While this challenge has been approached using supervised, semi-supervised, and unsupervised methods, our approach falls into the category of unsupervised object discovery techniques \citep{greff2017nem, greff2019iodine, burgess2019monet, eslami2016attend, lin2020space, goyal2020object, crawford2019spatially, zorna2021parts, slot-attention, ke2021systematic, engelcke2019genesis, goyal2019recurrent, goyal2021neural}. These papers mainly consider synthetic datasets, relying on a set of vectors known as 'slots' to represent objects. They use various architectural priors and objectives to group image features into slots corresponding to distinct objects. Recent works \citep{singh2022illiterate, boqsa, seitzer2023bridging, choudhury2021unsupervised, Wang2022SelfSupervisedTF, wang2023objectcentric} have proposed improvements that scale these slot-based methods to real-world datasets. While most datasets utilize pixel-wise reconstruction objectives, \citep{seitzer2023bridging, wang2023objectcentric} are the only two works utilizing objectives in the latent space, akin to our proposed approach. 'Dinosaur,' introduced by \citep{seitzer2023bridging}, applies slot attention on features from a pretrained self-supervised encoder like DINO \citep{dino}, employing feature reconstruction as the training objective. 

\citep{wang2023objectcentric}  employs cycle consistency objective on features  from a pretrained encoder. The differences between our work and theirs is that - (1) they primarily consider pretrained and frozen encoder backbones, while we train the encoders in our model from scratch, (2) They use a fixed threshold $T$ as compared to the adaptive threshold which we use, and (3) They only use the cycle consistency objectives for training their models, while we use the consistency objectives as auxillary objectives while training models from scratch.

One caveat of using pretrained encoders such as DINO is that it limits the method's applicability to domains for which the pretrained encoders are trained. %Secondly, another difference between our works and theirs is that we evaluate the usefulness of the learned object-centric representations on two downstream RL tasks while they don't perform any downstream evaluations in their paper.  %have clustered features obtained from a pretrained model into slots. While many of these works have shown promising results for object discovery, they do not explicitly enforce consistency or correspondence between the image features and the slots. Most existing approaches rely on architectural priors or various objectives (e.g., contrastive objectives \citep{choudhury2021unsupervised, Henaff2022ObjectDA}) to recover objects. However, these approaches do not explicitly encode the information that each slot must represent a distinct object. In contrast, our proposed cycle consistency objectives explicitly optimize for this, i.e., the objectives enforce each slot to represent a distinct object. Furthermore, our proposed objectives do not require multiple views of the scene, as in \citep{choudhury2021unsupervised}, or multiple networks, as in \citep{Henaff2022ObjectDA}. The proposed objectives are simple and can be easily integrated into any slot-based object discovery method.%Cycle consistency is a concept in deep learning where a mapping between two domains is learned by enforcing consistency between the original input and the output of the mapping by composing the mapping in a cycle. In other words, given an input in domain A, the cycle consistency objective ensures that the output of the mapping to domain B and back to domain A is similar to the original input in domain A. Similarly, given an input in domain B, the cycle consistency objective ensures that the output of the mapping to domain A and back to domain B is similar to the original input in domain B. This is often used in tasks such as image-to-image translation, where the goal is to learn a mapping between two domains. Cycle consistency has been shown to be effective for learning mappings between different domains in deep learning.

%The idea of using transitivity as a way to regularize structured data has a long history. Invisual tracking, enforcing simple forward-backward consistency has been a standard trick for decades

\vspace{-1mm}
\paragraph{Cycle Consistency}  Cycle consistency is a concept in deep learning that enables the learning of a consistent mapping between two domains in cases where one-to-one data is not available. It relies on the transitivity property to enforce structure in representations and has been successfully used for learning good representations in images, videos, and language. Numerous studies \citep{wang2013coseg, wandg2014coseg, kule2013network, zach2010disambiguating, zhou2015flowweb, zhou2016learning, zhou2015multi, hoffman2018cycada, zhu2017unpaired} have employed cycle consistency in image-based tasks such as co-segmentation, domain transfer, and image matching. In these works, cycle consistency is typically used as an objective function that ensures the consistency of the mapping between the source and target domains, and the inverse mapping from the target domain back to the source domain. For example, in \citep{zhu2017unpaired}, the source and target domains consist of images from distinct styles. Cycle consistency has also been utilized as a self-supervised representation learning technique in videos. Various studies \citep{dwibedi2019temporal, wang2019unsupervised, li2019joint, lai2019self, jabri2020space, Hadji_2021_CVPR} have used the cycle consistency objective to enforce temporal consistency in videos, ensuring that there is a path forward from frame $i$ to frame $k$, and the path backward from frame $k$ lands back on frame $i$. Furthermore, cycle consistency has been applied in different language settings where paired training data is not available (\citep{Huang2020cycle, hu2020neural, lee-lee-2022-type}), and in multimodal visual-language settings \citep{chen2019canzsl, Wang2022CounterfactualCL, shah2019cycle}. Our work differs from previous works in that we apply the cycle consistency objective for object discovery. Additionally, the cycle consistency objective is applied to the latent space consisting of the slots and features in this case, whereas previous studies have primarily focused on applying the objective to the domains of language, images, or video.

\section{Experiments}
In our experiments, we initially assess the efficacy of our approach as a representation learning technique by implementing it in two distinct downstream RL tasks. Subsequently, we evaluate the performance of the proposed objectives in object discovery by applying them to a range of synthetic and real-world tasks. Our findings demonstrate that the proposed approach not only learns superior representations, resulting in strong performance in downstream RL tasks but also enhances object discovery and generalization across a variety of benchmarks.

\paragraph{Object-Centric Approaches} We consider multiple object-centric approaches that utilize the slot attention model \citep{slot-attention}. Our proposed objectives are agnostic to the object-centric approach used and only require that the underlying base object-centric approach employs slot attention. We consider the following as base object-centric approaches: (1) Slot Attention \citep{slot-attention}; (2) SLATE \citep{singh2022illiterate}; (3) Dinosaur \citep{seitzer2023bridging}; and (4) MoTok \citep{bao2023object}. We integrate the proposed cycle consistency objectives into these base approaches. Slot Attention employs a top-down iterative attention mechanism to discover slots from image features obtained using a convolutional encoder. The model is trained via a reconstruction loss, where the decoded images are obtained by a mixture-based decoder \citep{watters2019spatial}. SLATE uses slot attention to discover slots but substitutes the convolutional encoder in slot attention with a dVAE \citep{oord2017neural, dalle}, and the convolutional decoder with an auto-regressive transformer \citep{dalle}. Dinosaur also utilizes the slot attention module for discovering slots but replaces the convolutional encoder with a pretrained and fixed DINO encoder \citep{dino}. The decoder in Dinosaur reconstructs pretrained DINO features instead of reconstructing the image in pixel space. MoTok \citep{bao2023object} is an object-centric model for videos that includes a slot attention module and utilizes motion segmentation as auxiliary information for discovering objects. We refer to our objectives as the \textsc{cyclic} objectives.

\paragraph{Datasets and Environments} For our downstream RL tasks, we use the Atari and Causal World environments \citep{ahmed2020causalworld}. In Atari, we consider 13 games and report the mean returns across 10 episodes for each game, similar to \citep{chen2021decision}. In Causal World, we consider 2 variants of the object goal task, where the agent should move a robotic arm towards a target object, using success rate as the performance metric.

For object discovery, we consider both synthetic and real-world datasets. In the synthetic datasets, we utilize Shapestacks \citep{Groth2018ShapeStacksLV}, ObjectsRoom \citep{multiobjectdatasets19}, and ClevrTex \citep{Karazija2021ClevrTexAT}. We evaluate segmentation performance on these datasets using the Adjusted Rand Index (ARI) \citep{Hubert1985ComparingP} and reconstruction performance using the mean squared error (MSE). Specifically, we calculate FG-ARI for these datasets, which is the same as ARI but ignores background information.

Additionally, we consider the Movi-C and Movi-E datasets, which contain synthetically constructed 3D scans of realistic objects on realistic backgrounds \citep{Greff2022KubricAS}. Lastly, we evaluate the proposed method on multi-object segmentation using the COCO \citep{coco} and ScanNet \citep{dai2017scannet} datasets. For this task, we report the AP score \citep{Everingham2014ThePV}, precision score, and recall score."
\subsection{Representation Learning for Downstream RL tasks}

\paragraph{Atari}

One crucial aspect of any representation learning method is that the learned representations should prove useful in downstream tasks. In this section, we explore the usefulness of the proposed approach in the context of Atari games. \citep{chen2021decision} introduced the decision transformer model, which learns to play various games in the Atari suite by imitating suboptimal trajectories from a learned agent. More details about the decision transformer model are presented in Appendix Section \ref{sec:dt_appendix}. In this work, we modify the monolithic state representation used in the decision transformer to an object-centric one.

The monolithic state representation of an observation is a $D$-dimensional vector obtained by passing the Atari observations through a convolutional encoder. Each observation is a stack of 4 frames. To derive the corresponding object-centric version, we use the convolutional encoder and the slot attention module from \citep{slot-attention} to encode each observation. Consequently, each observation is encoded into $N$ slots instead of a single vector.

To ensure that the slots learn the correct object-centric representation, we augment the decision transformer loss with the slot attention loss: $\mathcal{L} = \mathcal{L}{DT} + \mathcal{L}{Reconstruction}$. Additionally, we also include the cycle consistency objectives in the loss: $\mathcal{L} = \mathcal{L}{DT} + \mathcal{L}{Reconstruction} + \lambda_{sfs'} \mathcal{L}{sfs'} + \lambda{fsf'} \mathcal{L}_{fsf'}$. We compare our method to the baseline decision transformer and an object-centric variant of the decision transformer (DT + SA), where we use the slot attention style reconstruction loss but omit the cycle consistency objectives.

\begin{wraptable}{r}{0.5\textwidth}
\scriptsize
\vspace{-8mm}
\renewcommand{\arraystretch}{1.2}
\setlength{\tabcolsep}{0.8pt}
\centering
\caption{\textbf{Atari}. Here we present mean returns across 10 episodes on various games from the Atari suite. Results averaged across 5 seeds.}
\vspace{1mm}
\begin{tabular}{|c|c|c|c|}
\hline
Game & DT & DT + SA & DT + SA + Cyclic \\
\hline
Pong & \g{11.0}{5.727} & \g{7.4}{6.184} & \highlight{\g{14.8}{2.482}} \\
\hline
Breakout & \g{70.6}{20.539} & \g{93.4}{24.121} & \highlight{\g{110.2}{11.107}} \\
\hline
Seaquest & \highlight{\g{1172.4}{175.779}} & \g{444.0}{179.738} & \g{663.2}{111.014} \\
\hline
Qbert & \g{5485.2}{1995.256} & \g{5275.2}{862.894} & \highlight{\g{7393.8}{1982.989}} \\
\hline
Asterix & \g{523.333}{61.146} & \g{471.667}{253.388} & \highlight{\g{785.0}{153.677}} \\
\hline
Assault & \g{387.333}{23.099} & \g{430.667}{83.003} & \highlight{\g{462.0}{128.693}} \\
\hline
Boxing & \g{78.0}{1.633} & \g{77.333}{1.247} & \highlight{\g{78.667}{0.943}} \\
\hline
Carnival & \g{486.0}{343.872} & \g{814.0}{49.423} & \highlight{\g{836.667}{91.277}} \\
\hline
Freeway & \highlight{\g{26.667}{0.471}} & \g{21.0}{0.816} & \g{23.0}{0.816} \\
\hline
Crazy Climber & \g{76564.0}{24713.859} & \g{51490.0}{28676.178} & \highlight{\g{94254.0}{7569.641}} \\
\hline
BankHeist & \g{11.4}{6.974} & \g{105.0}{88.808} & \highlight{\g{144.8}{116.68}} \\
\hline
Space Invaders & \highlight{\g{602.2}{67.972}} & \g{392.0}{189.67} & \g{598.2}{52.147} \\
\hline
MsPacman & \g{1461.4}{329.76} & \g{1220.8}{237.301} & \highlight{\g{1900.0}{206.364}} \\
\hline
\end{tabular}
\label{tab:dt}
\vspace{-12mm}
\end{wraptable}
The performance comparison in Table \ref{tab:dt} reveals that the decision transformer, when augmented solely with object-centric representations obtained from slot attention (DT + SA), exhibits competitive performance across most games compared to the original decision transformer (DT). However, when the object-centric decision transformer is combined with the proposed cycle consistency objectives, it surpasses the baseline decision transformer in 10 out of 13 games. This outcome highlights the significance of the proposed cycle consistency objective in learning robust object-centric representations capable of performing well in downstream tasks.

\paragraph{Causal World} We consider the object goal task from the causal world environment \citep{ahmed2020causalworld}, following the same setup as \citep{yoon2023investigation}. In the Object Goal task, the agent controls a tri-finger robot placed in a bowl containing a target object and several distractor objects. The task for the agent is to move the tri-finger robot toward the target object without touching the distractors.

\begin{wrapfigure}{r}{0.5\textwidth}
\vspace{-6mm}
\centering
\includegraphics[width = \linewidth, scale = 0.3]{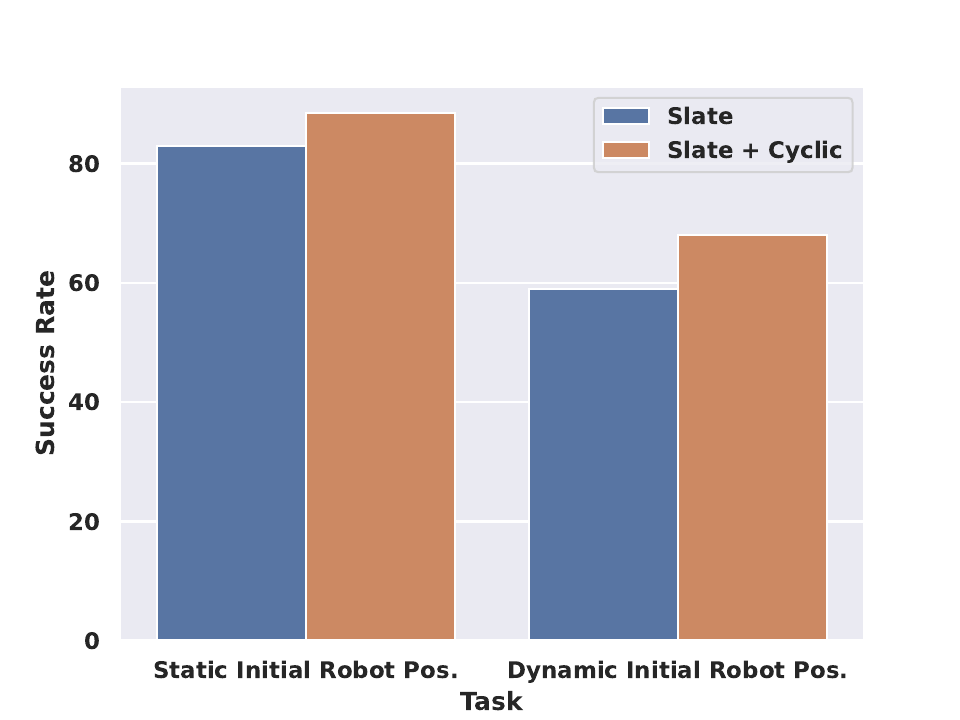}
\caption{\textbf{Causal World RL}: In this figure, we present results for both causal world tasks. We can observe that Slate models pretrained with the cyclic objectives achieve a superior success rate compared to the baseline Slate Model. Results averaged across 5 seeds.} \label{fig:cw_results}
\vspace{-6mm}
\end{wrapfigure}

Similar to \citep{yoon2023investigation}, we first pretrain a Slate model augmented with the cycle consistency objectives on random rollouts from the object goal task. The baseline for this task is a Slate model pretrained on the same dataset of random rollouts. We visualize the learned object masks in Figure \ref{fig:cw_vis}. It's evident that the proposed objectives enable the model to learn more accurate object masks compared to the baseline.

Next, we train a policy using proximal policy optimization (PPO) \citep{schulman2017ppo}. We employ a transformer-based policy network with inputs as slot representations from the object-centric models along with a CLS token. The resulting output corresponding to the CLS token is used to generate the action distribution and the value. The object-centric model remains frozen at this stage.

We consider two variants of the object-goal task: (1) Static Initial Robot Position - where the initial position of the tri-finger robot remains the same across episodes; (2) Dynamic Initial Robot Position - where the initial position of the tri-finger robot varies across episodes. Results for this task are presented in Figure \ref{fig:cw_results}. In both cases, models pretrained with the cyclic objective outperform those without it. More details about the pretraining procedure and the policy are provided in Appendix Section \ref{sec:cw}.

\subsection{Object Discovery}
\paragraph{Synthetic Datasets} We proceed to analyze the object discovery performance of the proposed approach. Initially, we focus on synthetic tasks, adhering to the setup outlined in \citep{dittadi2022generalization}. We enhance the slot-attention auto-encoder \citep{slot-attention} model with the cycle consistency objectives. Slot attention utilizes a convolutional encoder to acquire image features and a mixture-based decoder for reconstruction \citep{watters2019spatial}. The encoder outputs features $\tilde{F} \in \mathbb{R}^{t \times t \times D}$, where $t = 64$. Prior to applying the cycle consistency objectives, we downsample the features to $\tilde{t} = 16$, then project them using an MLP to match the dimensionality of the slots. Additionally, we normalize the slots using their L2 norm before incorporating the cycle consistency objectives. Further details concerning the architecture and hyperparameters are available in the Appendix.

%\begin{table}[h]
%    \centering
%    \footnotesize
%    \renewcommand{\arraystretch}{1.3}
%\setlength{\tabcolsep}{1.7pt}
 %       \caption{\textbf{ClevrTex Generalization}. Here we present transfer results on clevrtex. We use an improved version of slot attention called BO-Slot Attention \citep{boqsa} as the baseline. We train the model on the \textit{Full} split and transfer it to the \textit{Camo} and \textit{OOD} splits. We find that the proposed approach outperforms the baseline on both the transfer splits. Results averaged across 3 seeds.}

 %   \begin{tabular}{|c|c|c|c|}
 %        \hline
 %        & ClevrTex-Full & ClevrTex-CAMO & ClevrTex-OOD \\
 %        \hline
 %        Model & ARI-FG $\uparrow$ & ARI-FG $\uparrow$ &  ARI-FG $\uparrow$ \\
 %        \hline
  %       BO-Slot Attention \citep{boqsa} & \g{0.8016}{0.0227} & \g{0.7028}{0.0236} & \g{0.7157}{0.0224} \\
  %       \hline
   %      + \textsc{Cyclic} & \highlight{\g{0.8131}{0.0118}} & \highlight{\g{0.7049}{0.011}} & \highlight{\g{0.7357}{0.0187}} \\
   %      \hline
   % \end{tabular}
   % \label{tab:synthetic_clevrtex}
%\end{table}

\begin{table}[h]
\centering
\scriptsize
\renewcommand{\arraystretch}{1.3}
\setlength{\tabcolsep}{1.5pt}

\caption{\textbf{Synthetic Datasets Segmentation}. In this table we compare the slot attention model augmented with the proposed \textsc{Cyclic} objectives against the original slot attention model \citep{slot-attention}. As shown in the table, we observe that the proposed objectives result in performance gains across all the considered datasets. Results averaged across 3 seeds.}\label{tab:synthetic_1}
\begin{tabular}{|c | c | c |  c| c | c | c | c | c |} 
\hline
 &  &  &  \multicolumn{2}{c|}{ObjectsRoom} & \multicolumn{2}{c|}{ShapeStacks} & \multicolumn{2}{c|}{ClevrTex} \\
\hline
 Model & $\lambda_{sfs'}$ &  $\lambda_{fsf'}$ & MSE $\downarrow$ & FG-ARI $\uparrow$ & MSE $\downarrow $ & FG-ARI $\uparrow$ & MSE $\downarrow $ & FG-ARI $\uparrow$ \\
 \hline
 Slot-Attention & 0 & 0 & \g{0.0018}{0.0004} &  \g{0.7819}{0.08} & \g{0.004}{0.0004} & \g{0.7738}{0.05} & \g{0.007}{0.001}  & \g{0.6240}{0.223} \tablefootnote{
 We took the result for SA from \citep{Karazija2021ClevrTexAT}, as we  were unable to reproduce the same result in a statistically consistent manner. With our implementation, we got  FG-ARI score of \g{0.5864}{0.01}. To be fair, we have reported the score from the original paper \citep{Karazija2021ClevrTexAT}.} \\ 
 \hline
 \hline
  + \textsc{Cyclic} & $> 0$ & 0 & \g{0.0019}{0.0003} & \g{0.7832}{0.05} & \g{0.004}{0.0010}   & \g{0.5491}{0.4521} & \g{0.007}{0.0001}
 & \g{0.6640}{0.05} \\
 \hline
   + \textsc{Cyclic} & 0 & $ > 0$ & \g{0.0015}{0.0002}  & \g{0.8120}{0.06}  & \g{0.004}{0.0003}  & \g{0.7755}{0.06} & \g{0.007}{0.0001}
 & \g{0.4974}{0.03} \\
 \hline
   + \textsc{Cyclic} & $ > 0 $ & $ > 0 $ & \highlight{\g{0.0015}{0.0002}} & \highlight{\g{0.8169}{0.03}} & \highlight{\g{0.0037}{0.0001}}  & \highlight{\g{0.7838}{0.02}} & \g{0.007}{0.0001}
 & \highlight{ \g{0.7245}{0.01}} \\
 \hline
\end{tabular}
\vspace{-2mm}
\end{table}

\paragraph{Results.}Table \ref{tab:synthetic_1} presents the quantitative results for all three datasets. It's evident that augmenting Slot Attention with the proposed objectives results in improved factorization (measured by FG-ARI) and superior reconstruction (measured by MSE). Additionally, we note that while the inclusion of only one of the objectives doesn't notably impact reconstruction performance, having both objectives proves crucial for achieving robust factorization.
\vspace{-3mm}

\begin{wrapfigure}{r}{0.5\textwidth}
\vspace{-12mm}
  \begin{center}
    \includegraphics[width=0.48\textwidth]{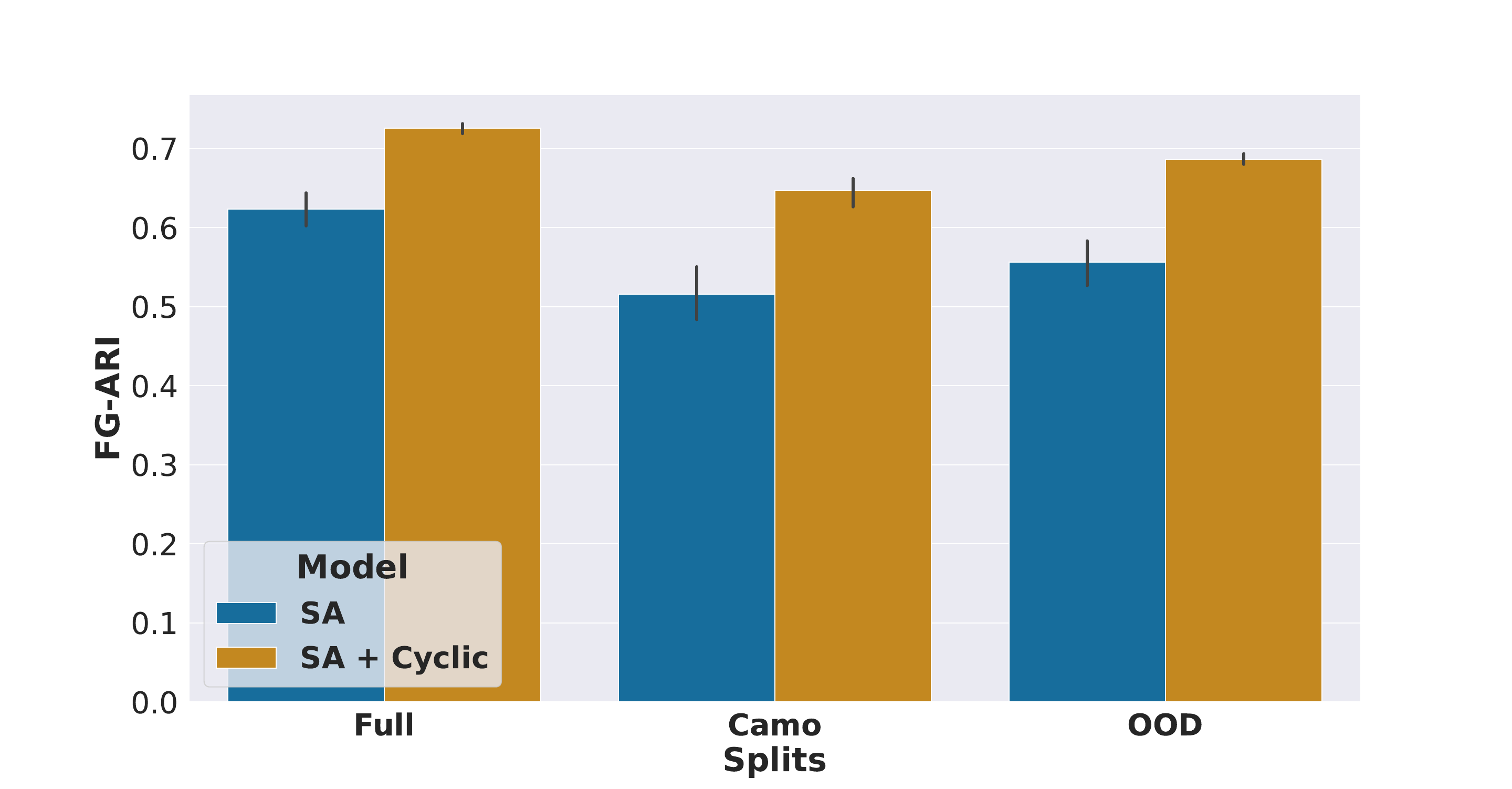}
  \end{center}
  \caption{\textbf{ClevrTex Generalization}. Here we present transfer results on clevrtex.  We find that the proposed approach outperforms the baseline on both the transfer splits. Results averaged across 5 seeds.}
  \label{fig:clevrtex_transfer}
  \vspace{-4mm}
\end{wrapfigure}
\paragraph{ClevrTex Generalization.}
The ClevrTex dataset offers two generalization splits that enable us to assess the generalization capabilities of the proposed approach: (1) CAMO - comprising scenes where specific objects are camouflaged, and (2) OOD - utilizing 25 new materials unseen during training. We train the models using the complete training set of ClevrTex and then apply them to both splits. Figure \ref{fig:clevrtex_transfer} displays the results, illustrating that the proposed method consistently outperforms the baseline on these generalization splits.

% \begin{table}%{r}{0.8\linewidth}
% \scriptsize
%   \centering
%   \vspace{-3mm}
    
%     \begin{tabular}{| c| c |}
%         \hline
%         Model & FG-ARI \\
%         \hline
%          SA + Cyclic (last iteration only) &  \g{0.5453}{0.06} \\
%          \hline
%          SA + Cyclic (all iterations)&  \highlight{\g{0.7245}{0.01}} \\
%          \hline
%     \end{tabular} 
%     \caption{\textbf{Iteration Ablation} We observe that the performance suffers a significant drop when the cycle consistency objecives are only applied to the last iteration of slot attention. Results averaged across 5 seeds.}
%     \label{tab:ablation_iteration}
%     \vspace{-5mm}
% \end{table}

\begin{table}[h]
\vspace{2mm}
  \begin{minipage}{.5\linewidth}
    \centering
    \scriptsize
     \begin{tabular}{|c|c|}
        \hline
        \textbf{Model} & \textbf{FG-ARI} \\
        \hline
        MoTok & 67.38 \\
        \hline
        MoTok + Cyclic & \highlight{72.48} \\
        \hline
    \end{tabular}
    \vspace{3mm}
    \caption{\textbf{Cycle Consistency Objectives with Motion Guidance}. Here we incorporate the cycle consistency objectives into the MoTok model presented in \citep{bao2023object}. We find that incorporating the proposed objectives into MoTok results in improved performance on the considered video-based object discovery task. We consider the Movi-E dataset for this experiment. } \label{tab:motok_results}
  \end{minipage}
  \hspace{1mm}
  \begin{minipage}{.5\linewidth}
    \centering
    \tiny
    \renewcommand{\arraystretch}{1.2}
\setlength{\tabcolsep}{0.8pt}
     \begin{tabular}{|l|c|c|c|c|}
    \hline
    Model & \multicolumn{2}{c|}{MOVi-C} & \multicolumn{2}{c|}{MOVi-E} \\
    \hline
    & FG-ARI & mBO & FG-ARI & mBO \\
    \hline
    DINOSAUR (ViT-B/8) & \g{68.9}{0.4} & \g{38.0}{0.2} & \g{65.1}{1.2} & \g{33.5}{0.1} \\
    \hline
    + \textsc{Cyclic}  & \highlight{\g{72.4}{2.1}} & \highlight{\g{40.2}{0.5}} & \highlight{\g{69.7}{1.6}} & \highlight{\g{37.2}{0.4}} \\
    \hline
  \end{tabular}
  \vspace{3mm}
    \caption{\textbf{Cycle Consistency Objectives with Pretrained Encoders} In this table we demonstrate the improvements achieved by the cyclic objectives when added to the DINOSAUR model from \citep{seitzer2023bridging} based on the pretrained ViT-B/8 encoder. We can see that proposed method achieves superior performance on both the datasets.} \label{tab:movi-comparison}
  \end{minipage}
  
\end{table}

% \begin{wraptable}{r}{0.4\textwidth}
%     \centering
%     \caption{\textbf{Cycle Consistency Objectives with motion guidance}. Here we incorporate the cycle consistency objectives into the MoTok model presented in \citep{bao2023object}. We find that incorporating the proposed objectives into MoTok results in improved performance on the considered video-based object discovery task. We consider the Movi-E dataset for this experiment. }
%     \begin{tabular}{|c|c|}
%         \hline
%         \textbf{Model} & \textbf{FG-ARI} \\
%         \hline
%         MoTok & 67.38 \\
%         MoTok + Cyclic & \highlight{72.48} \\
%         \hline
%     \end{tabular}
%     \label{tab:motok_results}
%     \vspace{-10mm}
% \end{wraptable}

% Your content he

\vspace{-5mm}
\paragraph{Cycle Consistency Objective with Object Discovery Approaches That Utilize Motion Guidance}
Several object discovery approaches utilize additional information, such as motion guidance, for discovering objects \citep{bao2023object, karazija2022unsupervised}. These studies focus on object discovery within video settings. We aim to assess whether the proposed cycle consistency objectives contribute in this setting. To do so, we integrate the cycle consistency objectives into the slot attention module of the MoTok model introduced in \citep{bao2023object}. The results of this experiment are presented in Table \ref{tab:motok_results}. Our findings reveal that augmenting MoTok with the cycle consistency objectives leads to improved performance on the Movi-E dataset compared to the MoTok baseline.

% \begin{wraptable}{r}{4cm}
% \tiny
%   \centering
%   \vspace{-12mm}
%   \caption{\textbf{Cycle Consistency Objectives with Pretrained Encoders} In this table we demonstrate the improvements achieved by the cyclic objectives when added to the DINOSAUR model from \citep{seitzer2023bridging} based on the pretrained ViT-B/8 encoder. We can see that proposed method achieves superior performance on both the datasets.}
%   \vspace{1mm}
%   \begin{tabular}{|l|c|c|c|c|}
%     \hline
%     Model & \multicolumn{2}{c|}{MOVi-C} & \multicolumn{2}{c|}{MOVi-E} \\
%     \hline
%     & FG-ARI & mBO & FG-ARI & mBO \\
%     \hline
%     DINOSAUR (ViT-B/8) & \g{68.9}{0.4} & \g{38.0}{0.2} & \g{65.1}{1.2} & \g{33.5}{0.1} \\
%     + \textsc{Cyclic}  & \highlight{\g{72.4}{2.1}} & \highlight{\g{40.2}{0.5}} & \highlight{\g{69.7}{1.6}} & \highlight{\g{37.2}{0.4}} \\
%     \hline
%   \end{tabular}
%   \vspace{-12mm}
%   \label{tab:movi-comparison}
% \end{wraptable}
\paragraph{Cycle Consistency Objective with Pretrained Encoders}
 We utilize the setup presented in \citep{seitzer2023bridging}, which introduces the DINOSAUR model for object discovery. DINOSAUR employs slot attention on features obtained from a ViT-B/8 encoder pretrained using the methodology in \citep{dino}. The training objective involves a reconstruction loss to learn the slot attention parameters, with the features from the pretrained ViT model as targets for this reconstruction objective. For our proposed method, we introduce the cyclic objectives to the DINOSAUR model, applying them exclusively to the slots obtained from the last iteration of slot attention. In this case, we do not employ an EMA encoder, as the ViT image encoder remains frozen throughout training. This comparison is conducted on the MOVI datasets \citep{Greff2022KubricAS}. The results in Table \ref{tab:movi-comparison} demonstrate that DINOSAUR augmented with cyclic objectives outperforms the base DINOSAUR model on both datasets. This further underscores that the proposed cycle consistency objectives are agnostic to the underlying object discovery approach and only necessitate the use of a slot attention module.

\begin{figure}%{r}{0.6\textwidth}
  \centering
\vspace{-8mm}
  \subfigure[$\lambda_{sfs'}$ variation for $\lambda_{fsf'} = 0.01$]{
    \includegraphics[width=0.4\linewidth]{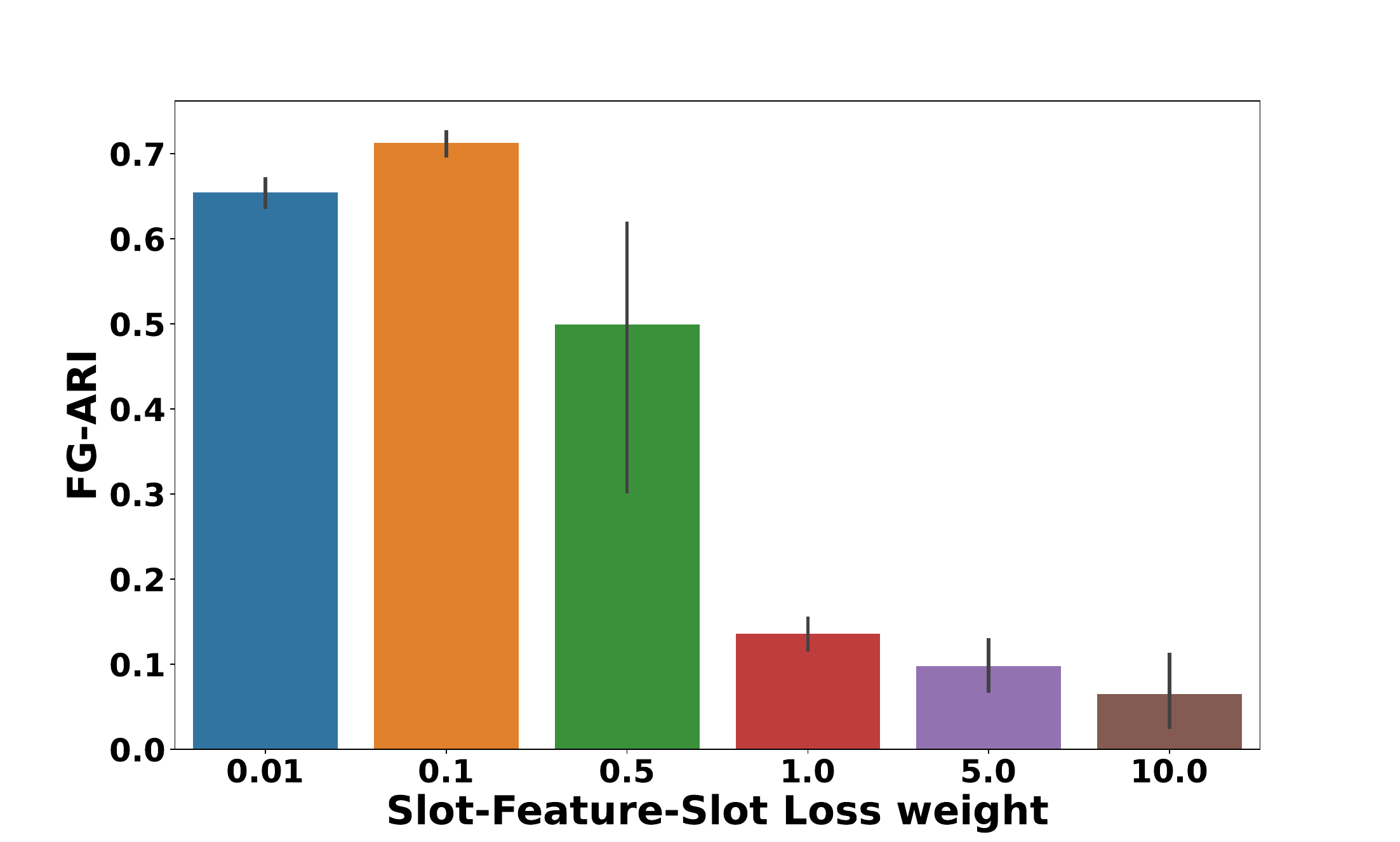} % Replace with your image file
    \label{fig:subfig1}
  }
  \subfigure[$\lambda_{fsf'}$ variations for $\lambda_{sfs'} = 0.1$]{
    \includegraphics[width=0.4\linewidth]{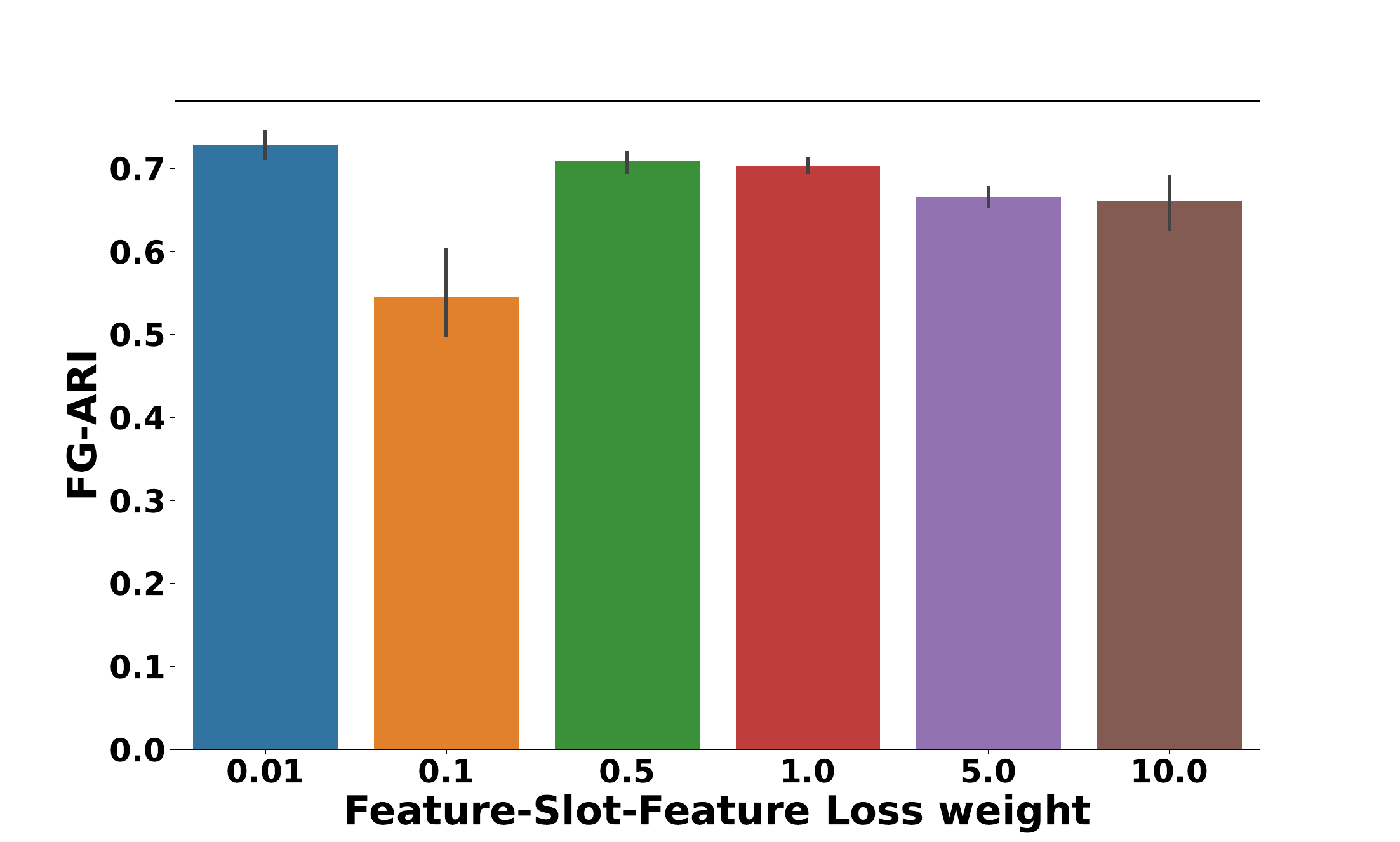} % Replace with your image file
    \label{fig:subfig2}
  }

  \caption{\textbf{Effect of Loss Coefficients }(a) In this we vary the \textsc{Slot-Feature-Slot} weight ($\lambda_{sfs'}$) and keep the \textsc{Feature-Slot-Feature} weight ($\lambda_{fsf'}$) fixed to 0.01. We can see that the model reaches peak performance for $\lambda_{sfs'} = 0.1$ and the performance degrades rapidly for higher values. (b) In this we vary $\lambda_{fsf'}$ and keep $\lambda_{sfs'}$ fixed to 0.1. We can see that the performance is much more stable here as compared to varying $\lambda_{sfs'}$.}
  \label{fig:loss_coef}
  \vspace{-5mm}
\end{figure}

\paragraph{Effect of Loss Coefficients} We examine the impact of the loss coefficients ($\lambda_{sfs'}$ and $\lambda_{fsf'}$) on the ClevrTex dataset in Figure \ref{fig:loss_coef}. Observing the variation in $\lambda_{sfs'}$ (Figure \ref{fig:loss_coef}(a)), we note a rapid performance degradation for higher values. This decline might be due to the \textsc{Slot-Feature-Slot} consistency becoming trivially satisfied if all features map to a single slot. Consequently, a high $\lambda_{sfs'}$ could bias the model towards such a trivial solution, leading to performance deterioration. In Figure \ref{fig:loss_coef}(b), the performance variation is notably more stable with varying values of $\lambda_{fsf'}$. This suggests that the model is relatively unaffected by changes in $\lambda_{fsf'}$.

\paragraph{Effect of Applying Objectives on All Iterations of Slot Attention}
One implementation detail of the proposed method involves applying the cycle consistency objectives to slots from all iterations of slot attention. We conduct an ablation study on this design choice by comparing it against a model where the cycle consistency objectives are solely applied to the last iteration of slot attention. The results are presented in Table \ref{tab:ablation_iteration}. A noticeable performance drop is observed when the cycle consistency objectives are exclusively applied to the last iteration, underscoring the importance of applying the objectives across all iterations.

\paragraph{Effect of EMA Encoder}
Another crucial aspect of the proposed approach involves using an EMA encoder to compute the feature-slot-feature supervision matrix $\tilde{\mathcal{F}}$. To evaluate the significance of the EMA encoder, we compare the performance of the proposed approach on the Shapestacks datasets with and without it. The comparison results are detailed in Table \ref{tab:ablation_ema}. It's evident that the model's performance significantly improves when employing the EMA encoder, highlighting its importance in the proposed approach.

\begin{table}[h]
\vspace{2mm}
  \begin{minipage}{.5\linewidth}
    \centering
    \scriptsize
     \begin{tabular}{| c| c |}
        \hline
        Model & FG-ARI \\
        \hline
         SA + Cyclic (last iteration only) &  \g{0.5453}{0.06} \\
         \hline
         SA + Cyclic (all iterations)&  \highlight{\g{0.7245}{0.01}} \\
         \hline
    \end{tabular} 
    \vspace{2mm}
    \caption{\textbf{Iteration Ablation} We observe that the performance suffers a significant drop when the cycle consistency objecives are only applied to the last iteration of slot attention. We run this experiment on the ClevrTex dataset.} \label{tab:ablation_iteration}
  \end{minipage}
  \hspace{1mm}
  \begin{minipage}{.5\linewidth}
    \centering
    \scriptsize
     \begin{tabular}{| c| c |}
        \hline
        Model & FG-ARI \\
        \hline
         SA + Cyclic (No EMA Encoder) &  \g{0.7028}{0.03} \\
         \hline
         SA + Cyclic (EMA Encoder)&  \highlight{\g{0.7838}{0.02}} \\
         \hline
    \end{tabular}
    \vspace{2mm}
    \caption{\textbf{EMA Encoder Ablation} We can see that the performance of the model without the EMA encoder is much worse than with it thus showing the importance of the EMA encoder. We run this experiment on the shapestacks dataset.} \label{tab:ablation_ema}
  \end{minipage}
  
\end{table}
\begin{table}[h]
\tiny
\renewcommand{\arraystretch}{1}
\setlength{\tabcolsep}{1pt}
    \centering
    \caption{\textbf{Real World Datasets Segmentation}. Here we present results for multi-object segmentation on real-world datasets. We augment the improved Slate model presented in \citep{boqsa} with the proposed cyclic objectives. We can see that the proposed approach outperforms all baselines on all metrics on both datasets. Results averaged across 3 seeds.}
   
    \begin{tabular}{| c | c | c | c | c || c| c | c | c |}
    \hline

         & \multicolumn{4}{c||}{ COCO} & \multicolumn{4}{c|}{Scannet} \\
         \hline
         Model & AP@05 $\uparrow$ & PQ $\uparrow$ & Precision $\uparrow$ & Recall $\uparrow$ &AP@05 $\uparrow$ & PQ $\uparrow$ & Precision $\uparrow$ & Recall $\uparrow$ \\
         \hline
         MONet \citep{burgess2019monet} & \g{11.8}{2.0} & \g{12.5}{1.1} & \g{16.1}{0.9} & \g{21.9}{1.7} & \g{24.8}{1.6} & \g{24.6}{1.6} & \g{31.0}{1.6} & \g{40.7}{1.8} \\
         \hline
         IODINE \citep{greff2019iodine} & \g{4.0}{1.2} & \g{6.3}{1.2} & \g{9.9}{1.8} & \g{10.8}{2.0} & \g{10.1}{2.9} & \g{13.7}{2.7} & \g{18.6}{4.2} & \g{24.4}{3.8} \\ 
         \hline
         Slot Attention \citep{slot-attention} & \g{0.8}{0.3} & \g{3.5}{1.2} & \g{5.3}{1.7} & \g{7.3}{2.2} & \g{5.7}{0.3} & \g{9.0}{1.5} & \g{12.4}{2.5} & \g{18.3}{2.7} \\
         \hline
         Implicit Slot Attention \citep{chang2023object} & \g{12.8}{4.8} & \g{13.7}{4.5} & \g{20.4}{6.0} & \g{24.6}{7.3} & \g{21.4}{6.8} & \g{23.4}{1.5} & \g{29.1}{7.8} & \g{34.5}{7.0} \\
         \hline
         BO-Slate \citep{boqsa} & \g{16.64}{1.00} & \g{17.48}{0.9}  & \g{25.49}{1.2} & \g{31.13}{1.5} &  \g{24.67}{3.2} & \g{23.55}{0.3} & \g{34.03}{0.4}& \g{38.74}{0.6} \\ 
         \hline
         BO-Slate + \textsc{Cyclic} & \highlight{\g{18.96}{0.9}} & \highlight{\g{18.81}{0.8}} & \highlight{\g{27.50}{1.4}} & \highlight{\g{33.20}{1.6}} & \highlight{\g{29.20}{1.1}} & \highlight{\g{26.09}{1.4}} & \highlight{\g{37.03}{1.4}} &  \highlight{\g{42.09}{1.8}} \\ 
          \hline
         % sam & 21.35 & 38.52 & 43.25 & 47.64 & 15.09 & 34.32 & 35.54 & 41.29 \\  
          %\hline
    \end{tabular}
    
    \label{tab:real_world_2}
\end{table}

\paragraph{Real-World Datasets}
For these experiments, we utilize the \textsc{BO-Slate} model \citep{boqsa} as our base model. \textsc{BO-Slate} represents an improved iteration of Slate \citep{singh2022illiterate}, primarily enhanced through learnable slot initializations. In our proposed method, we integrate the cycle consistency objectives into the \textsc{BO-Slate} model, denoted as BO-Slate + Cyclic. Further details are provided in the Appendix.

\paragraph{Multi-Object Segmentation}
The results for this task are presented in Table \ref{tab:real_world_2}. We employ the same dataset versions utilized in \citep{boqsa}. Our comparison involves the proposed approach against various object-centric models documented in the literature. Remarkably, the BO-Slate model, augmented with the proposed objectives, outperforms all the baseline models across all metrics. Further details about these metrics are available in the appendix.

\section{Future Work and Limitations}

Considerable research efforts have been dedicated to learning object-centric representations. However, the evaluation of these representations has primarily focused on unsupervised segmentation performance using metrics like ARI or IoU. Unfortunately, there is a lack of studies demonstrating the practical utility of object-centric representations across diverse downstream tasks. To address this gap, our work takes a step forward by showcasing the effectiveness of our approach in the context of Atari and Causal World. While we do not address all limitations of object-centric methods on downstream tasks, more work is needed to demonstrate the effectiveness of these models on more complex and real-world downstream tasks.

Moving forward, our objective is to shift our focus towards developing object-centric representations that prove valuable in a wide array of downstream tasks, spanning reinforcement learning to visual tasks like visual question answering and captioning. We aim to explore the efficacy of cycle consistency objectives in learning such representations and study what is lacking in building more pervasive object-centric representations.

\section{Acknowledgements}
This research was made possible in part by compute resources, software, and technical support provided by Mila (mila.quebec). The authors extend their appreciation to Nanda Harishankar Krishna for assistance in refining Figure \ref{fig:method_demo}. Gratitude is also owed to Ayush Chakravarthy and Vedant Shah for their valuable contributions during brainstorming sessions. Additionally, the authors would like to thank Nanda Harishankar Krishna, Vedant Shah, and Jithendaraa Subramanian for their thoughtful reviews and suggestions to improve this manuscript.

{\small
\bibliographystyle{abbrvnat}
\bibliography{egbib}
}

\include{appendix}
\end{document}

%% file: appendix.tex
\section*{\Large Appendix}

\section{Synthetic Dataset Experiments}

\begin{figure}
     \centering
     \includegraphics[width = \linewidth]{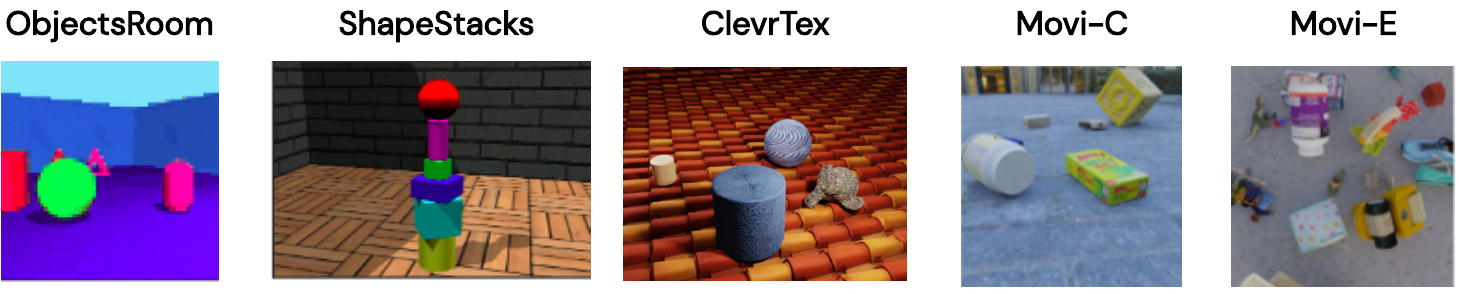}
     \caption{Here we show an example image from each synthetic dataset that we consider.}
     \label{fig:synthetic_dataset_demo}
 \end{figure}

 We consider the Shapestacks \citep{Groth2018ShapeStacksLV}, ObjectsRoom\citep{multiobjectdatasets19}, ClevrTex \citep{Karazija2021ClevrTexAT}, and MOVi datasets \citep{movi_dataset}. Figure \ref{fig:synthetic_dataset_demo} shows an example image from each dataset. 

\paragraph{Slot Attention Implementation Details} For ObjectsRoom, ShapeStacks, and ClevrTex, we use slot attention as our base model. Table \ref{tab:enc_dec_arch} shows the detailed architecture of the convolutional encoder and decoder used by the slot attention model.  Table \ref{tab:syn_hyper} indicates values of various hyperparameters used in these experiments.  For each experiment, we use 1 RTX8000 GPU. 

\begin{table}[]
 \tiny
 \renewcommand{\arraystretch}{1}
\setlength{\tabcolsep}{1.2pt}
    \centering
    \begin{tabular}{| c | c | c | c | c | c | c | c | c | c | c | c | c |} 
    \hline
     & \multicolumn{4}{|c|}{ObjectsRoom} & \multicolumn{4}{|c|}{ShapeStacks} & \multicolumn{4}{|c|}{ClevrTex} \\
    
    \hline
  Layer  & Channels & Kernel Size & Padding & Stride & Channels & Kernel Size & Padding & Stride & Channels & Kernel Size & Padding & Stride \\
    \hline
    \multicolumn{13}{|c|}{Convolutional Encoder} \\
    
    \hline
 Conv   & 32 & 5 & 2 & 1 & 32 & 5 & 2 & 1 & 64 & 5 & 2 & 1 \\
  Conv  & 32 & 5 & 2 & 1 & 32 & 5 & 2 & 1 & 64 & 5 & 2 & 1 \\
  Conv  & 32 & 5 & 2 & 1 & 32 & 5 & 2 & 1 & 64 & 5 & 2 & 1 \\
  Conv  & 32 & 5 & 2 & 1 & 32 & 5 & 2 & 1 & 64 & 5 & 2 & 1 \\
    
    \hline
    \multicolumn{13}{|c|}{Convolutional Decoder} \\
    \hline
    Conv. Trans & 32 & 5 & 2 & 1 & 32 & 5 & 2 & 1 & 64 & 5 & 2 & 2 \\
    Conv. Trans & 32 & 5 & 2 & 1 & 32 & 5 & 2 & 1 & 64 & 5 & 2 & 2 \\
    Conv. Trans & 32 & 5 & 2 & 1 & 32 & 5 & 2 & 1 & 64 & 5 & 2 & 2 \\
    Conv. Trans & 4 & 5 & 1 & 1 & 32 & 5 & 2 & 1 & 4 & 5 & 2 & 1 \\
    \hline

    \end{tabular}
    \vspace{1mm}
    \caption{Detailed architecture for the encoder and decoder used by the slot attention model used in the synthetic dataset experiments. Note that we use relu activations after every layer except the last layer.}
    \label{tab:enc_dec_arch}
\end{table}

\begin{table}[]
    \centering
    \begin{tabular}{| c |c | c | c |}
         \hline
         & ObjectsRoom & ShapeStacks & ClevrTex \\
         \hline
         Num. Slots & 7 & 7 & 11 \\
         Num. Iter & 3 & 3 & 3 \\
         Slot size & 64 & 64 & 64 \\
         MLP size & 128 & 128 & 128 \\
         Batch Size & 64 & 64 & 64 \\
         Optimizer & Adam & Adam & Adam \\
         LR & 0.0004 & 0.0004 & 0.0004 \\
         Total steps & 500k & 500k & 500k \\
         Warmup Steps & 10k & 10k & 5k \\
         Decay Steps & 100k & 100k & 50k \\
         $\lambda_{sfs'}$ & 0.1 & 0.1 & 0.1 \\
         $\lambda_{fsf'}$ & 0.01 & 0.01 & 0.01 \\
         $\tau$ & 0.1 & 0.1 & 0.1 \\
         $\tau_{sfs'}$ & 1 & 1 & 1 \\
         $\tau_{fsf'}$ & 0.01 & 0.01 & 0.01 \\
         $\theta_i$ & 0.8 & 0.8 & 0.8 \\
         Downsampled feature size & $16 \times 16$ & $16 \times 16$ & $32 \times 32$ \\
         EMA Decay rate & 0.995 & 0.995 & 0.995 \\
         \hline
    \end{tabular}
    \vspace{1mm}
    \caption{This table indicates all the values for various hyperparameters used in the synthetic dataset experiments.}
    \label{tab:syn_hyper}
\end{table}

\begin{table}[]
    \centering
    \begin{tabular}{|c|c|c | c|}
    \hline
    & ObjectsRoom & ShapeStacks  & ClevrTex \\
    \hline
    Model & FG-ARI & FG-ARI & FG-ARI \\
    \hline
         SA + Cyclic ($\mathcal{L}_{fsf'} \quad \forall \quad i, j$)& \g{0.7341}{0.07}& \g{0.7161}{0.01} &  \g{0.6349}{0.11}\\
         \hline
         SA + Cyclic ($\mathcal{L}_{fsf'} \quad \forall \quad i = j$)& \highlight{\g{0.8169}{0.03}}  & \highlight{\g{0.7838}{0.02}} & \highlight{\g{0.7245}{0.01}}\\  
    \hline
    \end{tabular}
    \vspace{1mm}
    \caption{Here we compare the performance of the model where $\mathcal{L}_{fsf'}$ is computed for all i, j to the model which computes $\mathcal{L}_{fsf'}$ for all i = j. We can see that the latter performs better than former thus showing the importance of computing the $\mathcal{L}_{fsf'}$ only for i = j. We perform this ablation on the ObjectsRoom, ShapeStacks, and ClevrTex datasets.}
    \label{tab:lfsf_ablation}
\end{table}

\paragraph{Ablation on $\mathcal{L}_{fsf'}$ application} In our case, we calculate $\mathcal{L}{fsf'}$ as follows: $\mathcal{L}{fsf'} = -p^{i \rightarrow j}\log(P(f_j \mid f_i)) \quad \forall \quad i = j$. Therefore, instead of computing the loss for all pairs of $i$ and $j$, we only compute it for the cases where $i$ is equal to $j$. This design choice is made because the supervision signal, $p^{i \rightarrow j}$, is a function of the pairwise feature similarity values. Obtaining accurate pairwise feature similarity values for all $i, j$ in a model trained from scratch is challenging. Hence, we limit the loss calculation to only the diagonal elements of the matrix, where $i = j$. To assess the significance of this design choice, we compare the performance of the proposed model with a model that computes $\mathcal{L}_{fsf'}$ for all $i, j$. The results of this study are presented in Table \ref{tab:lfsf_ablation}. Notably, computing $\mathcal{L}_{fsf'}$ solely for $i = j$ yields considerably better performance compared to computing it for all $i, j$.

\paragraph{Dinosaur Implementation Details} We use DINOSAUR \citep{seitzer2023bridging} as the base model to which we augment the cycle consistency objectives for our experiments on the MOVi-E and MOVi-C datasets. We use a pretrained ViT-B/8 model pretrained using the approach presented in DINO \citep{dino}. We use 10 slots for experiments on the MOVi-C dataset and 23 slots for experiments on the MOVi-E dataset. For both datasets, we use 3 slot attention iterations. In this case, we apply the cycle consistency objectives on the slots obtained from the last iteration of slot attention. We set $\lambda_{sfs'}$ to 5 and $\lambda_{fsf'}$ to 1. We use Adam optimizer with a learning rate of 4e-4. We run each experiment on 1 RTX8000 GPU.

\begin{figure}
    \centering
    \includegraphics[width = 0.7\linewidth, trim = {0 0 8cm 0}, clip]{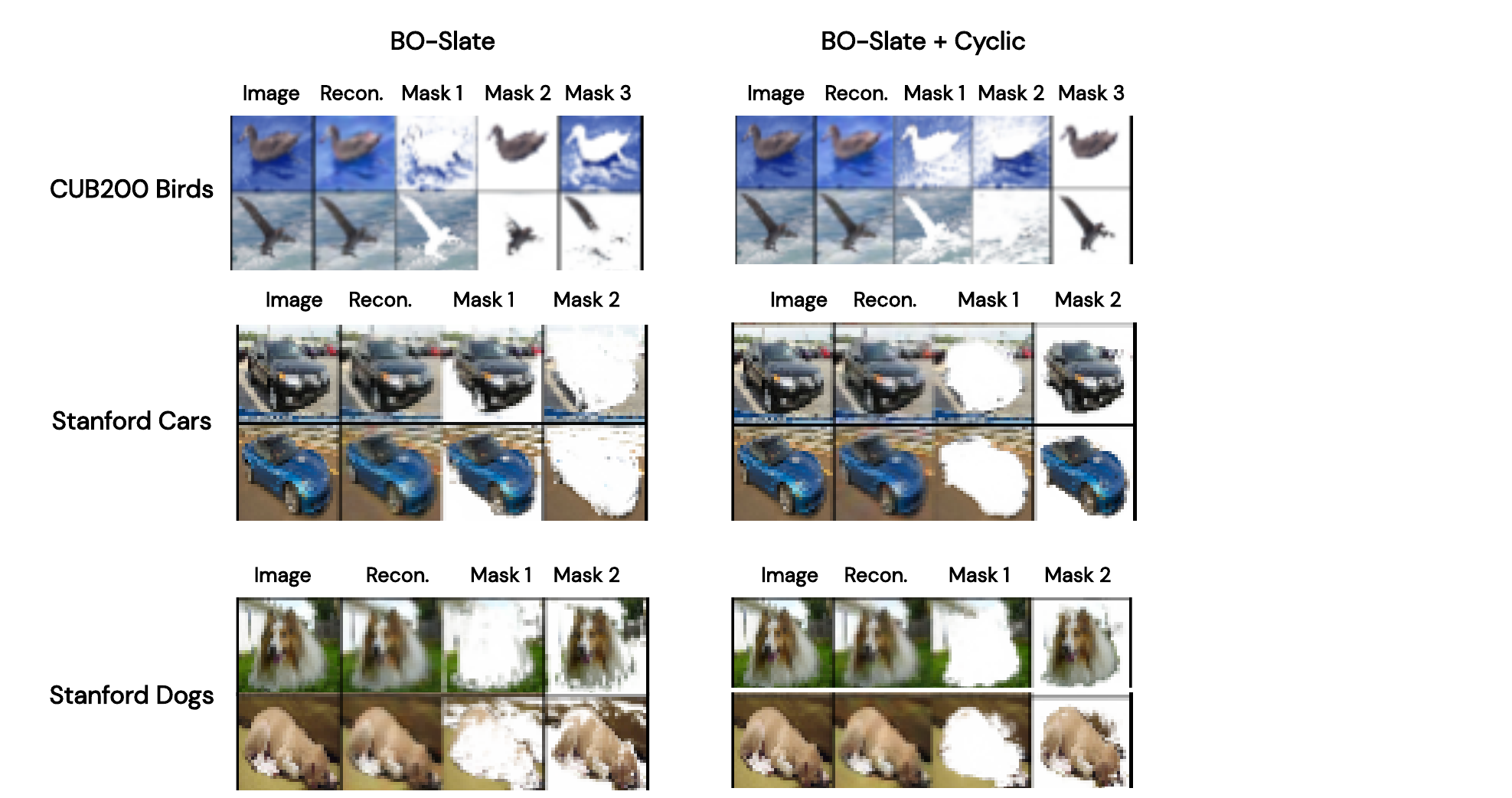}
    \caption{\textbf{Foreground Extraction Visualization}. This figure showcases the reconstruction and segmentation masks produced by the baseline model (BO-Slate) and the proposed cyclic model. Notably, we observe that the baseline model tends to mix the foreground with parts of the background, particularly for the Stanford Dogs and Stanford Cars datasets. In contrast, the use of our cyclic objectives leads to a significantly clearer separation of the foreground and background, resulting in a more accurate and refined representation of the objects of interest.}
    \label{fig:real_world_demo}
\end{figure}
\vspace{-6mm}

\section{Real World Dataset Experiments}
For these experiments, we use the BO-Slate \citep{boqsa} as the base model to which we augment the cycle consistency objectives. The values of all hyperparameters used for these experiments are shown in Table \ref{tab:slate_hyperparams}.

\begin{table}[]
    \centering
    \begin{tabular}{|c|c |c|}
    \hline
         \multirow{5}{*}{Training} & batch size & 64 \\
         & warmup steps & 10000 \\
         & learning rate & 1e-4 \\
         & max steps & 250k \\
         & Input image size & $96 \times 96$ \\
         \hline
         \multirow{4}{*}{dVAE} & vocabulary size & 1024 \\
         & Gumbel-Softmax annealing range & 1.0 to 0.1 \\
         & Gumbel-Softmax annealing steps & 30000 \\
         & lr-dVAE (no warmup) & 3e-4 \\
         \hline
         \multirow{4}{*}{ Transformer Decoder} & layer & 4\\
         & heads & 4 \\
         & dropout & 0.1 \\
         & hidden dimension & 256 \\
         \hline
         \multirow{3}{*}{Slot Attention Module} & slot dimension & 256 \\
         & iterations & 3 \\
         & $\sigma$ annealing steps & 30000 \\
         \hline
         \multirow{8}{*}{Cycle Consistency Objective} & $\lambda_{sfs'}$ & 10 \\
         & $\lambda_{fsf'}$ & 1 \\
         & $\tau$ & 0.1 \\
         &  $\tau_{sfs'}$ & 1\\
         & $\tau_{fsf'}$ & 0.01 \\
         & $\theta_i$ & 0.8 \\
         & EMA Decay rate & 0.900 \\
         & Downsampled Feature size & $24 \times 24$\\
         \hline
         
    \end{tabular}
    \vspace{1mm}
    \caption{This table shows various hyperparameters used in the real-world dataset experiments where we use BO-Slate as the base model.}
    \label{tab:slate_hyperparams}
\end{table}

\paragraph{Unsupervised Foreground Extraction} For foreground extraction, we use the Stanford dogs dataset \citep{dogs}, Stanford cars dataset \citep{cars}, CUB200 Birds dataset \citep{birds}, and flowers dataset \citep{flowers}. We evaluate the performance using the IoU (Intersection over union) and the Dice metrics. IoU is calculated by dividing the overlapping area between the ground-truth and predicted masks by the union area of the ground-truth and predicted masks. Dice is calculated as twice the area of overlap between the ground-truth and predicted mask divide by the combined number of pixels between the ground-truth and predicted masks. 

We use the \textsc{BO-Slate} model \citep{boqsa} as our base model. \textsc{BO-Slate} is an improved version of Slate \citep{singh2022illiterate} where the main improvements come from having learnable slot initializations. We apply the cycle consistency objectives as auxillary objectives to BO-Slate. %We apply the objective to slots obtained from all iterations of slot attention. We also downsample the features before applying the objective. The resolution of the downsampled features varies for each dataset. 

 The results for this task are presented in Table \ref{tab:real_world_1}. We observe that the proposed objectives helps the model achieve a superior performance compared to the baseline on all datasets. Additionally, in Figure \ref{fig:real_world_demo}, we visualize the reconstruction and segmentation masks from the model. We note that in certain cases, the baseline model tends to mix foreground and background information, whereas the same model augmented with the cyclic objectives is able to segregate the foreground and background information perfectly.

 \begin{table}[h]
\scriptsize	
\renewcommand{\arraystretch}{1}
\setlength{\tabcolsep}{1.2pt}
    \centering
    
    \caption{\textbf{Unsupervised Foreground Extraction}. Here we present results for unsupervised foreground extraction on the Stanford Dogs, Stanford Cars, and CUB 200 birds dataset. We augment our cyclic objectives to the improved version of Slate \citep{singh2022illiterate} presented in \citep{boqsa}. We can see that the performance of the Slate model improves when augmented with the proposed objectives thus showing the efficacy of our approach. Results averaged across 3 seeds.}
    \begin{tabular}{| c | c | c | c | c | c | c | c | c |}
    \hline
    
         & \multicolumn{2}{c |}{ Dogs } & \multicolumn{2}{c |}{ Cars } & \multicolumn{2}{c |}{Birds} & \multicolumn{2}{c |}{Flowers} \\
         \hline
         Model & IoU $\uparrow$ & Dice $\uparrow$ & IoU $\uparrow$ & Dice $\uparrow$ & IoU $\uparrow$ & Dice $\uparrow$ & IoU $\uparrow$ & Dice $\uparrow$ \\
         \hline
         BO-Slate & \g{0.7875}{0.05}  & \g{0.6681}{0.06} & \g{0.7686}{0.10} & \g{0.8647}{0.07} & \g{0.6129}{0.05} & \g{0.7473}{0.05} & \g{0.7461}{0.03} & \g{0.8340}{0.02} \\ 
         \hline
          + \textsc{Cyclic} & \highlight{\g{0.8456}{0.04}}& \highlight{\g{0.7462}{0.06}} & \highlight{\g{0.8447}{0.02}} & \highlight{\g{0.9145}{0.02}} & \highlight{\g{0.6497}{0.01}} & \highlight{\g{0.7797}{0.009}} & \highlight{\g{0.7745}{0.01}} & \highlight{\g{0.8525}{0.01}} \\ 
          \hline
    \end{tabular}
    
    \label{tab:real_world_1}
\end{table}

We follow the hyperparameters mentioned in Table \ref{tab:slate_hyperparams}. We use 2 slots for all foreground extraction experiments except for Birds for which we use 3 slots. To downsample the features obtained from the encoder for computing the cycle consistency objectives, we use a 2 layered convolutional network in which each layer has kernel size 4, stride 2, and padding 1. We use a relu activation between the two layers. We run each experiment on 1 RTX8000 GPU. 

\paragraph{Multi-Object Segmentation} For multi-object segmentation, we use the coco \citep{coco} and scannet \citep{dai2017scannet} datasets. We use the following metrics to evaluate performance - 
\begin{itemize}
    \item AP@05: AP is a metric used in object detection to measure the accuracy and relevance of detection results based on precision and recall values.
    \item Panoptic Quality (PQ): PQ is a comprehensive metric for evaluating object segmentation that combines segmentation quality and instance-level recognition performance into a single score.
    \item Precision score:  Precision measures the proportion of correctly predicted foreground pixels among all the pixels predicted as foreground, indicating the accuracy of the segmentation results.
    \item Recall Score: Recall measures the proportion of correctly predicted foreground pixels among all the ground truth foreground pixels, indicating the completeness or coverage of the segmentation results.
\end{itemize}

We use the same hyperparameters as presented in Table \ref{tab:slate_hyperparams}. For calculating the cycle consistency objective, we downsample the features output by the encoder using a similar convolutional network as used in the Foreground Extraction task.

\section{Object Centric Models in Atari}\label{sec:dt_appendix}
We follow the exact setup from decision transformer \citep{chen2021decision} for this experiment. Decision Transformer models the offline RL problem as a conditional sequence modelling task. This is done by feeding into the model the states, actions, and the return-to-go $\hat{R}_c = \sum_{c'=c}^C r_c$, where $c$ denotes the timesteps. This results in the following trajectory representation: $\tau = \big( \hat{R}_1, s_1, a_1, \hat{R}_2, s_2, a_2, \hat{R}_3, s_3, a_3, \ldots \big)$, where $a_c$ denotes the actions and $s_c$ denotes the states.  At test time, the start state $s_1$ and desired return $\hat{R}_1$ is fed into the model and it autoregressively generates the rest of the trajectory.

The original state representations $s_i$ are $D$-dimensional vectors obtained by passing the atari observations through a convolutional encoder. Note that each observation is a stack of 4 frames. To obtain the corresponding object-centric version of this, we use the convolutional encoder and the slot attention module from \citep{slot-attention} to encode each observation. Therefore, each observation is encoded into $N$ slots resulting in a decision transformer trajectory - $\tau = \big( \hat{R}_1, \{s_1^1, s_1^2 \ldots, s_1^N\}, a_1, \hat{R}_2, \{s_2^1, s_2^2 \ldots, s_2^N\}, a_2, \hat{R}_3, \{s_3^1, s_3^2 \ldots, s_3^N\}, a_3, \ldots \big)$. 

As mentioned in the main text, we augment the action-prediction loss from decision transformer with the reconstruction loss from slot attention and the proposed cycle consistency objectives. 

We use a 6-layered transformer and 8 attention heads with an embedding size 128. We use a batch size of 64. We use a context length of 50 for Pong and a context length of 30 for Seaquest, Breakout, and Qbert. We keep all other hyperparameters same as mentioned in \citep{chen2021decision}. For the slot attention implementation, we define the architecture of the encoder and decoder in Table \ref{tab:dt_conv}. The values of the other hyperparameters related to the slot attention module and the cycle consistency objectives are presented in Table \ref{tab:dt_hparams}. The input image size for the atari experiments is $84 \times 84$. We downsample the features to size $21 \times 21$ to compute the cycle consistency objectives. We use a 2-layered convolutional network for this downsampling where each layer has kernel size 4, stride 2, and padding 1. 

\begin{table}[]
\scriptsize
    \centering
    \begin{tabular}{|c | |c | c | c | c | c| c |}
    \hline 
    & Channels & Layer & Kernel Size & Padding & Stride & output padding \\
    \hline
         \multirow{4}{*}{Convolutional Encoder} & Conv. & 64 & 5 & 2 & 1 & \\
         & Conv. &  64 & 5 & 2 & 1 & \\
         & Conv. &  64 & 5 & 2 & 1 & \\
         & Conv & 64 & 5 & 2 & 1 & \\
    \hline
    \multirow{4}{*}{Convolutional Decoder} & Conv Trans. & 64 & 7 & 0 & 2 & 0 \\
    & Conv Trans. & 32 & 3 & 1 & 2 & 1 \\
    & Conv Trans. & 5 (4 for frames + 1 for mask) & 3 & 1 & 2 & 1 \\
    \hline
    \end{tabular}
    \vspace{1mm}
    \caption{Architecture of the encoder and decoder used in the slot attention module for the decision transformer model. }
    \label{tab:dt_conv}
\end{table}

\begin{table}[]
    \centering
    \begin{tabular}{|c|c|}
    \hline
    Hyperparameter & Value \\
    \hline
       slot dimension & 256 \\
         iterations & 3 \\
         \hline
         $\lambda_{sfs'}$ & 0.1 \\
         $\lambda_{fsf'}$ & 0.01 \\
         $\tau$ & 0.1 \\
         $\tau_{sfs'}$ & 1\\
         $\tau_{fsf'}$ & 0.01 \\
          $\theta_i$ & 0.8 \\
          EMA Decay rate & 0.995 \\
          Downsampled Feature size & $21 \times 21$\\
          \hline
    \end{tabular}
    \vspace{1mm}
    \caption{Here we present the values for the various hyperparameters used in the slot attention module for the decision transformer experiments.}
    \label{tab:dt_hparams}
\end{table}

\section{Object Centric Models in Causal World} \label{sec:cw}
We follow the same setup as \citep{yoon2023investigation} for this experiment. We first petrain the object centric model on 1000000 trajectories from the object goal task of the causal world environment. The baseline object centric model in our case is Slate. We augment it with the proposed cycle consistency objectives for our model. Both the object-centric models utilize 6 slots. We train them for 200k steps.  

For training the policy, we use PPO. The agent is a transformer-based model which takes the slots as input along with a CLS token and outputs a distribution over actions and a value. We train the agent for 1000000 interaction steps. 

%% file: main.bbl
\begin{thebibliography}{71}
\providecommand{\natexlab}[1]{#1}
\providecommand{\url}[1]{\texttt{#1}}
\expandafter\ifx\csname urlstyle\endcsname\relax
  \providecommand{\doi}[1]{doi: #1}\else
  \providecommand{\doi}{doi: \begingroup \urlstyle{rm}\Url}\fi

\bibitem[Ahmed et~al.(2020)Ahmed, Träuble, Goyal, Neitz, Wüthrich, Bengio,
  Schölkopf, and Bauer]{ahmed2020causalworld}
O.~Ahmed, F.~Träuble, A.~Goyal, A.~Neitz, M.~Wüthrich, Y.~Bengio,
  B.~Schölkopf, and S.~Bauer.
\newblock Causalworld: A robotic manipulation benchmark for causal structure
  and transfer learning, 2020.

\bibitem[Bao et~al.(2023)Bao, Tokmakov, Wang, Gaidon, and
  Hebert]{bao2023object}
Z.~Bao, P.~Tokmakov, Y.-X. Wang, A.~Gaidon, and M.~Hebert.
\newblock Object discovery from motion-guided tokens, 2023.

\bibitem[Burgess et~al.(2019)Burgess, Matthey, Watters, Kabra, Higgins,
  Botvinick, and Lerchner]{burgess2019monet}
C.~P. Burgess, L.~Matthey, N.~Watters, R.~Kabra, I.~Higgins, M.~M. Botvinick,
  and A.~Lerchner.
\newblock Monet: Unsupervised scene decomposition and representation.
\newblock \emph{CoRR}, abs/1901.11390, 2019.
\newblock URL \url{http://arxiv.org/abs/1901.11390}.

\bibitem[Caron et~al.(2021)Caron, Touvron, Misra, J{\'{e}}gou, Mairal,
  Bojanowski, and Joulin]{dino}
M.~Caron, H.~Touvron, I.~Misra, H.~J{\'{e}}gou, J.~Mairal, P.~Bojanowski, and
  A.~Joulin.
\newblock Emerging properties in self-supervised vision transformers.
\newblock \emph{CoRR}, abs/2104.14294, 2021.
\newblock URL \url{https://arxiv.org/abs/2104.14294}.

\bibitem[Chang et~al.(2023)Chang, Griffiths, and Levine]{chang2023object}
M.~Chang, T.~L. Griffiths, and S.~Levine.
\newblock Object representations as fixed points: Training iterative refinement
  algorithms with implicit differentiation, 2023.

\bibitem[Chen et~al.(2021)Chen, Lu, Rajeswaran, Lee, Grover, Laskin, Abbeel,
  Srinivas, and Mordatch]{chen2021decision}
L.~Chen, K.~Lu, A.~Rajeswaran, K.~Lee, A.~Grover, M.~Laskin, P.~Abbeel,
  A.~Srinivas, and I.~Mordatch.
\newblock Decision transformer: Reinforcement learning via sequence modeling,
  2021.

\bibitem[Chen and He(2020)]{simsiam}
X.~Chen and K.~He.
\newblock Exploring simple siamese representation learning.
\newblock \emph{CoRR}, abs/2011.10566, 2020.
\newblock URL \url{https://arxiv.org/abs/2011.10566}.

\bibitem[Chen et~al.(2019)Chen, Li, Luo, Huang, and Yang]{chen2019canzsl}
Z.~Chen, J.~Li, Y.~Luo, Z.~Huang, and Y.~Yang.
\newblock {CANZSL:} cycle-consistent adversarial networks for zero-shot
  learning from natural language.
\newblock \emph{CoRR}, abs/1909.09822, 2019.
\newblock URL \url{http://arxiv.org/abs/1909.09822}.

\bibitem[Choudhury et~al.(2021)Choudhury, Laina, Rupprecht, and
  Vedaldi]{choudhury2021unsupervised}
S.~Choudhury, I.~Laina, C.~Rupprecht, and A.~Vedaldi.
\newblock Unsupervised part discovery from contrastive reconstruction.
\newblock \emph{CoRR}, abs/2111.06349, 2021.
\newblock URL \url{https://arxiv.org/abs/2111.06349}.

\bibitem[Crawford and Pineau(2019)]{crawford2019spatially}
E.~Crawford and J.~Pineau.
\newblock Spatially invariant unsupervised object detection with convolutional
  neural networks.
\newblock In \emph{Proceedings of the Thirty-Third AAAI Conference on
  Artificial Intelligence and Thirty-First Innovative Applications of
  Artificial Intelligence Conference and Ninth AAAI Symposium on Educational
  Advances in Artificial Intelligence}, AAAI'19/IAAI'19/EAAI'19. AAAI Press,
  2019.
\newblock ISBN 978-1-57735-809-1.
\newblock \doi{10.1609/aaai.v33i01.33013412}.
\newblock URL \url{https://doi.org/10.1609/aaai.v33i01.33013412}.

\bibitem[Dai et~al.(2017)Dai, Chang, Savva, Halber, Funkhouser, and
  Nie{\ss}ner]{dai2017scannet}
A.~Dai, A.~X. Chang, M.~Savva, M.~Halber, T.~Funkhouser, and M.~Nie{\ss}ner.
\newblock Scannet: Richly-annotated 3d reconstructions of indoor scenes.
\newblock In \emph{Proc. Computer Vision and Pattern Recognition (CVPR), IEEE},
  2017.

\bibitem[Dittadi et~al.(2022)Dittadi, Papa, De~Vita, Sch{\"o}lkopf, Winther,
  and Locatello]{dittadi2022generalization}
A.~Dittadi, S.~Papa, M.~De~Vita, B.~Sch{\"o}lkopf, O.~Winther, and
  F.~Locatello.
\newblock Generalization and robustness implications in object-centric
  learning.
\newblock In \emph{International Conference on Machine Learning}, 2022.

\bibitem[Dwibedi et~al.(2019)Dwibedi, Aytar, Tompson, Sermanet, and
  Zisserman]{dwibedi2019temporal}
D.~Dwibedi, Y.~Aytar, J.~Tompson, P.~Sermanet, and A.~Zisserman.
\newblock Temporal cycle-consistency learning.
\newblock \emph{CoRR}, abs/1904.07846, 2019.
\newblock URL \url{http://arxiv.org/abs/1904.07846}.

\bibitem[Elsayed et~al.(2022)Elsayed, Mahendran, van Steenkiste, Greff, Mozer,
  and Kipf]{elsayed2022savi}
G.~F. Elsayed, A.~Mahendran, S.~van Steenkiste, K.~Greff, M.~C. Mozer, and
  T.~Kipf.
\newblock {SAV}i++: Towards end-to-end object-centric learning from real-world
  videos.
\newblock In A.~H. Oh, A.~Agarwal, D.~Belgrave, and K.~Cho, editors,
  \emph{Advances in Neural Information Processing Systems}, 2022.

\bibitem[Engelcke et~al.(2019)Engelcke, Kosiorek, Jones, and
  Posner]{engelcke2019genesis}
M.~Engelcke, A.~R. Kosiorek, O.~P. Jones, and I.~Posner.
\newblock Genesis: Generative scene inference and sampling with object-centric
  latent representations.
\newblock \emph{arXiv preprint arXiv:1907.13052}, 2019.

\bibitem[Eslami et~al.(2016)Eslami, Heess, Weber, Tassa, Szepesvari,
  kavukcuoglu, and Hinton]{eslami2016attend}
S.~M.~A. Eslami, N.~Heess, T.~Weber, Y.~Tassa, D.~Szepesvari, k.~kavukcuoglu,
  and G.~E. Hinton.
\newblock Attend, infer, repeat: Fast scene understanding with generative
  models.
\newblock In D.~Lee, M.~Sugiyama, U.~Luxburg, I.~Guyon, and R.~Garnett,
  editors, \emph{Advances in Neural Information Processing Systems}, volume~29.
  Curran Associates, Inc., 2016.

\bibitem[Everingham et~al.(2014)Everingham, Eslami, Gool, Williams, Winn, and
  Zisserman]{Everingham2014ThePV}
M.~Everingham, S.~M.~A. Eslami, L.~V. Gool, C.~K.~I. Williams, J.~M. Winn, and
  A.~Zisserman.
\newblock The pascal visual object classes challenge: A retrospective.
\newblock \emph{International Journal of Computer Vision}, 111:\penalty0
  98--136, 2014.

\bibitem[Ghorbani et~al.(2020)Ghorbani, Mahdaviani, Thaler, K{\"{o}}rding,
  Cook, Blohm, and Troje]{movi_dataset}
S.~Ghorbani, K.~Mahdaviani, A.~Thaler, K.~P. K{\"{o}}rding, D.~J. Cook,
  G.~Blohm, and N.~F. Troje.
\newblock Movi: {A} large multipurpose motion and video dataset.
\newblock \emph{CoRR}, abs/2003.01888, 2020.
\newblock URL \url{https://arxiv.org/abs/2003.01888}.

\bibitem[Goyal et~al.(2019)Goyal, Lamb, Hoffmann, Sodhani, Levine, Bengio, and
  Sch{\"o}lkopf]{goyal2019recurrent}
A.~Goyal, A.~Lamb, J.~Hoffmann, S.~Sodhani, S.~Levine, Y.~Bengio, and
  B.~Sch{\"o}lkopf.
\newblock Recurrent independent mechanisms.
\newblock \emph{arXiv preprint arXiv:1909.10893}, 2019.

\bibitem[Goyal et~al.(2020)Goyal, Lamb, Gampa, Beaudoin, Levine, Blundell,
  Bengio, and Mozer]{goyal2020object}
A.~Goyal, A.~Lamb, P.~Gampa, P.~Beaudoin, S.~Levine, C.~Blundell, Y.~Bengio,
  and M.~Mozer.
\newblock Object files and schemata: Factorizing declarative and procedural
  knowledge in dynamical systems.
\newblock \emph{arXiv preprint arXiv:2006.16225}, 2020.

\bibitem[Goyal et~al.(2021{\natexlab{a}})Goyal, Didolkar, Lamb, Badola, Ke,
  Rahaman, Binas, Blundell, Mozer, and Bengio]{goyal2021coordination}
A.~Goyal, A.~Didolkar, A.~Lamb, K.~Badola, N.~R. Ke, N.~Rahaman, J.~Binas,
  C.~Blundell, M.~Mozer, and Y.~Bengio.
\newblock Coordination among neural modules through a shared global workspace.
\newblock \emph{arXiv preprint arXiv:2103.01197}, 2021{\natexlab{a}}.

\bibitem[Goyal et~al.(2021{\natexlab{b}})Goyal, Didolkar, Ke, Blundell,
  Beaudoin, Heess, Mozer, and Bengio]{goyal2021neural}
A.~Goyal, A.~R. Didolkar, N.~R. Ke, C.~Blundell, P.~Beaudoin, N.~Heess, M.~C.
  Mozer, and Y.~Bengio.
\newblock Neural production systems.
\newblock In A.~Beygelzimer, Y.~Dauphin, P.~Liang, and J.~W. Vaughan, editors,
  \emph{Advances in Neural Information Processing Systems}, 2021{\natexlab{b}}.

\bibitem[Greff et~al.(2017)Greff, van Steenkiste, and
  Schmidhuber]{greff2017nem}
K.~Greff, S.~van Steenkiste, and J.~Schmidhuber.
\newblock Neural expectation maximization.
\newblock \emph{CoRR}, abs/1708.03498, 2017.
\newblock URL \url{http://arxiv.org/abs/1708.03498}.

\bibitem[Greff et~al.(2019)Greff, Kaufman, Kabra, Watters, Burgess, Zoran,
  Matthey, Botvinick, and Lerchner]{greff2019iodine}
K.~Greff, R.~L. Kaufman, R.~Kabra, N.~Watters, C.~Burgess, D.~Zoran,
  L.~Matthey, M.~M. Botvinick, and A.~Lerchner.
\newblock Multi-object representation learning with iterative variational
  inference.
\newblock \emph{CoRR}, abs/1903.00450, 2019.
\newblock URL \url{http://arxiv.org/abs/1903.00450}.

\bibitem[Greff et~al.(2022)Greff, Belletti, Beyer, Doersch, Du, Duckworth,
  Fleet, Gnanapragasam, Golemo, Herrmann, Kipf, Kundu, Lagun, Laradji, Liu,
  Meyer, Miao, Nowrouzezahrai, Oztireli, Pot, Radwan, Rebain, Sabour, Sajjadi,
  Sela, Sitzmann, Stone, Sun, Vora, Wang, Wu, Yi, Zhong, and
  Tagliasacchi]{Greff2022KubricAS}
K.~Greff, F.~Belletti, L.~Beyer, C.~Doersch, Y.~Du, D.~Duckworth, D.~J. Fleet,
  D.~Gnanapragasam, F.~Golemo, C.~Herrmann, T.~Kipf, A.~Kundu, D.~Lagun, I.~H.
  Laradji, H.-T. Liu, H.~Meyer, Y.~Miao, D.~Nowrouzezahrai, C.~Oztireli,
  E.~Pot, N.~Radwan, D.~Rebain, S.~Sabour, M.~S.~M. Sajjadi, M.~Sela,
  V.~Sitzmann, A.~Stone, D.~Sun, S.~Vora, Z.~Wang, T.~Wu, K.~M. Yi, F.~Zhong,
  and A.~Tagliasacchi.
\newblock Kubric: A scalable dataset generator.
\newblock \emph{2022 IEEE/CVF Conference on Computer Vision and Pattern
  Recognition (CVPR)}, pages 3739--3751, 2022.

\bibitem[Grill et~al.(2020)Grill, Strub, Altch\'{e}, Tallec, Richemond,
  Buchatskaya, Doersch, Avila~Pires, Guo, Gheshlaghi~Azar, Piot, kavukcuoglu,
  Munos, and Valko]{byol}
J.-B. Grill, F.~Strub, F.~Altch\'{e}, C.~Tallec, P.~Richemond, E.~Buchatskaya,
  C.~Doersch, B.~Avila~Pires, Z.~Guo, M.~Gheshlaghi~Azar, B.~Piot,
  k.~kavukcuoglu, R.~Munos, and M.~Valko.
\newblock Bootstrap your own latent - a new approach to self-supervised
  learning.
\newblock In H.~Larochelle, M.~Ranzato, R.~Hadsell, M.~Balcan, and H.~Lin,
  editors, \emph{Advances in Neural Information Processing Systems}, volume~33,
  pages 21271--21284. Curran Associates, Inc., 2020.

\bibitem[Groth et~al.(2018)Groth, Fuchs, Posner, and
  Vedaldi]{Groth2018ShapeStacksLV}
O.~Groth, F.~B. Fuchs, I.~Posner, and A.~Vedaldi.
\newblock Shapestacks: Learning vision-based physical intuition for generalised
  object stacking.
\newblock \emph{ArXiv}, abs/1804.08018, 2018.

\bibitem[Hadji et~al.(2021)Hadji, Derpanis, and Jepson]{Hadji_2021_CVPR}
I.~Hadji, K.~G. Derpanis, and A.~D. Jepson.
\newblock Representation learning via global temporal alignment and
  cycle-consistency.
\newblock In \emph{Proceedings of the IEEE/CVF Conference on Computer Vision
  and Pattern Recognition (CVPR)}, pages 11068--11077, June 2021.

\bibitem[He et~al.(2019)He, Fan, Wu, Xie, and Girshick]{moco}
K.~He, H.~Fan, Y.~Wu, S.~Xie, and R.~B. Girshick.
\newblock Momentum contrast for unsupervised visual representation learning.
\newblock \emph{CoRR}, abs/1911.05722, 2019.
\newblock URL \url{http://arxiv.org/abs/1911.05722}.

\bibitem[Hoffman et~al.(2018)Hoffman, Tzeng, Park, Zhu, Isola, Saenko, Efros,
  and Darrell]{hoffman2018cycada}
J.~Hoffman, E.~Tzeng, T.~Park, J.-Y. Zhu, P.~Isola, K.~Saenko, A.~Efros, and
  T.~Darrell.
\newblock {C}y{CADA}: Cycle-consistent adversarial domain adaptation.
\newblock In J.~Dy and A.~Krause, editors, \emph{Proceedings of the 35th
  International Conference on Machine Learning}, volume~80 of \emph{Proceedings
  of Machine Learning Research}, pages 1989--1998. PMLR, 10--15 Jul 2018.
\newblock URL \url{https://proceedings.mlr.press/v80/hoffman18a.html}.

\bibitem[Hu et~al.(2020)Hu, Wang, Zhou, and Xiong]{hu2020neural}
X.~Hu, R.~Wang, D.~Zhou, and Y.~Xiong.
\newblock Neural topic modeling with cycle-consistent adversarial training.
\newblock \emph{CoRR}, abs/2009.13971, 2020.
\newblock URL \url{https://arxiv.org/abs/2009.13971}.

\bibitem[Huang et~al.(2020)Huang, Zhu, Xiong, Zhang, Hu, and
  Xu]{Huang2020cycle}
Y.~Huang, W.~Zhu, D.~Xiong, Y.~Zhang, C.~Hu, and F.~Xu.
\newblock Cycle-consistent adversarial autoencoders for unsupervised text style
  transfer.
\newblock \emph{CoRR}, abs/2010.00735, 2020.
\newblock URL \url{https://arxiv.org/abs/2010.00735}.

\bibitem[Hubert and Arabie(1985)]{Hubert1985ComparingP}
L.~J. Hubert and P.~Arabie.
\newblock Comparing partitions.
\newblock \emph{Journal of Classification}, 2:\penalty0 193--218, 1985.

\bibitem[Jabri et~al.(2020)Jabri, Owens, and Efros]{jabri2020space}
A.~Jabri, A.~Owens, and A.~A. Efros.
\newblock Space-time correspondence as a contrastive random walk.
\newblock \emph{CoRR}, abs/2006.14613, 2020.
\newblock URL \url{https://arxiv.org/abs/2006.14613}.

\bibitem[Jia et~al.(2022)Jia, Liu, and Huang]{boqsa}
B.~Jia, Y.~Liu, and S.~Huang.
\newblock Improving object-centric learning with query optimization, 2022.
\newblock URL \url{https://arxiv.org/abs/2210.08990}.

\bibitem[Kabra et~al.(2019)Kabra, Burgess, Matthey, Kaufman, Greff, Reynolds,
  and Lerchner]{multiobjectdatasets19}
R.~Kabra, C.~Burgess, L.~Matthey, R.~L. Kaufman, K.~Greff, M.~Reynolds, and
  A.~Lerchner.
\newblock Multi-object datasets.
\newblock https://github.com/deepmind/multi-object-datasets/, 2019.

\bibitem[Karazija et~al.(2021)Karazija, Laina, and
  Rupprecht]{Karazija2021ClevrTexAT}
L.~Karazija, I.~Laina, and C.~Rupprecht.
\newblock Clevrtex: A texture-rich benchmark for unsupervised multi-object
  segmentation.
\newblock \emph{ArXiv}, abs/2111.10265, 2021.

\bibitem[Karazija et~al.(2022)Karazija, Choudhury, Laina, Rupprecht, and
  Vedaldi]{karazija2022unsupervised}
L.~Karazija, S.~Choudhury, I.~Laina, C.~Rupprecht, and A.~Vedaldi.
\newblock Unsupervised multi-object segmentation by predicting probable motion
  patterns, 2022.

\bibitem[Ke et~al.(2021)Ke, Didolkar, Mittal, Goyal, Lajoie, Bauer, Rezende,
  Bengio, Mozer, and Pal]{ke2021systematic}
N.~R. Ke, A.~Didolkar, S.~Mittal, A.~Goyal, G.~Lajoie, S.~Bauer, D.~Rezende,
  Y.~Bengio, M.~Mozer, and C.~Pal.
\newblock Systematic evaluation of causal discovery in visual model based
  reinforcement learning.
\newblock \emph{arXiv preprint arXiv:2107.00848}, 2021.

\bibitem[Khosla et~al.(2012)Khosla, Jayadevaprakash, Yao, and Fei-Fei]{dogs}
A.~Khosla, N.~Jayadevaprakash, B.~Yao, and L.~Fei-Fei.
\newblock Novel dataset for fine-grained image categorization : Stanford dogs.
\newblock 2012.

\bibitem[Kipf et~al.(2021)Kipf, Elsayed, Mahendran, Stone, Sabour, Heigold,
  Jonschkowski, Dosovitskiy, and Greff]{kipf2021conditional}
T.~Kipf, G.~F. Elsayed, A.~Mahendran, A.~Stone, S.~Sabour, G.~Heigold,
  R.~Jonschkowski, A.~Dosovitskiy, and K.~Greff.
\newblock Conditional object-centric learning from video.
\newblock \emph{CoRR}, abs/2111.12594, 2021.
\newblock URL \url{https://arxiv.org/abs/2111.12594}.

\bibitem[Krause et~al.(2013)Krause, Stark, Deng, and Fei-Fei]{cars}
J.~Krause, M.~Stark, J.~Deng, and L.~Fei-Fei.
\newblock 3d object representations for fine-grained categorization.
\newblock In \emph{4th International IEEE Workshop on 3D Representation and
  Recognition (3dRR-13)}, Sydney, Australia, 2013.

\bibitem[Lai and Xie(2019)]{lai2019self}
Z.~Lai and W.~Xie.
\newblock Self-supervised learning for video correspondence flow.
\newblock \emph{CoRR}, abs/1905.00875, 2019.
\newblock URL \url{http://arxiv.org/abs/1905.00875}.

\bibitem[Lee and Lee(2022)]{lee-lee-2022-type}
S.~Lee and M.~Lee.
\newblock Type-dependent prompt {C}ycle{QAG} : Cycle consistency for multi-hop
  question generation.
\newblock In \emph{Proceedings of the 29th International Conference on
  Computational Linguistics}, pages 6301--6314, Gyeongju, Republic of Korea,
  Oct. 2022. International Committee on Computational Linguistics.
\newblock URL \url{https://aclanthology.org/2022.coling-1.549}.

\bibitem[Li et~al.(2019)Li, Liu, Mello, Wang, Kautz, and Yang]{li2019joint}
X.~Li, S.~Liu, S.~D. Mello, X.~Wang, J.~Kautz, and M.~Yang.
\newblock Joint-task self-supervised learning for temporal correspondence.
\newblock \emph{CoRR}, abs/1909.11895, 2019.
\newblock URL \url{http://arxiv.org/abs/1909.11895}.

\bibitem[Lin et~al.(2014)Lin, Maire, Belongie, Hays, Perona, Ramanan,
  Doll{\'a}r, and Zitnick]{coco}
T.-Y. Lin, M.~Maire, S.~J. Belongie, J.~Hays, P.~Perona, D.~Ramanan,
  P.~Doll{\'a}r, and C.~L. Zitnick.
\newblock Microsoft coco: Common objects in context.
\newblock In \emph{European Conference on Computer Vision}, 2014.

\bibitem[Lin et~al.(2020)Lin, Wu, Peri, Sun, Singh, Deng, Jiang, and
  Ahn]{lin2020space}
Z.~Lin, Y.~Wu, S.~V. Peri, W.~Sun, G.~Singh, F.~Deng, J.~Jiang, and S.~Ahn.
\newblock {SPACE:} unsupervised object-oriented scene representation via
  spatial attention and decomposition.
\newblock \emph{CoRR}, abs/2001.02407, 2020.
\newblock URL \url{http://arxiv.org/abs/2001.02407}.

\bibitem[Locatello et~al.(2020)Locatello, Weissenborn, Unterthiner, Mahendran,
  Heigold, Uszkoreit, Dosovitskiy, and Kipf]{slot-attention}
F.~Locatello, D.~Weissenborn, T.~Unterthiner, A.~Mahendran, G.~Heigold,
  J.~Uszkoreit, A.~Dosovitskiy, and T.~Kipf.
\newblock Object-centric learning with slot attention.
\newblock In H.~Larochelle, M.~Ranzato, R.~Hadsell, M.~Balcan, and H.~Lin,
  editors, \emph{Advances in Neural Information Processing Systems}, volume~33,
  pages 11525--11538. Curran Associates, Inc., 2020.
\newblock URL
  \url{https://proceedings.neurips.cc/paper/2020/file/8511df98c02ab60aea1b2356c013bc0f-Paper.pdf}.

\bibitem[Nilsback and Zisserman(2006)]{flowers}
M.-E. Nilsback and A.~Zisserman.
\newblock A visual vocabulary for flower classification.
\newblock In \emph{IEEE Conference on Computer Vision and Pattern Recognition},
  volume~2, pages 1447--1454, 2006.

\bibitem[Ramesh et~al.(2021)Ramesh, Pavlov, Goh, Gray, Voss, Radford, Chen, and
  Sutskever]{dalle}
A.~Ramesh, M.~Pavlov, G.~Goh, S.~Gray, C.~Voss, A.~Radford, M.~Chen, and
  I.~Sutskever.
\newblock Zero-shot text-to-image generation.
\newblock \emph{CoRR}, abs/2102.12092, 2021.
\newblock URL \url{https://arxiv.org/abs/2102.12092}.

\bibitem[Schulman et~al.(2017)Schulman, Wolski, Dhariwal, Radford, and
  Klimov]{schulman2017ppo}
J.~Schulman, F.~Wolski, P.~Dhariwal, A.~Radford, and O.~Klimov.
\newblock Proximal policy optimization algorithms.
\newblock \emph{CoRR}, abs/1707.06347, 2017.
\newblock URL \url{http://arxiv.org/abs/1707.06347}.

\bibitem[Seitzer et~al.(2023)Seitzer, Horn, Zadaianchuk, Zietlow, Xiao,
  Simon-Gabriel, He, Zhang, Sch{\"o}lkopf, Brox, and
  Locatello]{seitzer2023bridging}
M.~Seitzer, M.~Horn, A.~Zadaianchuk, D.~Zietlow, T.~Xiao, C.-J. Simon-Gabriel,
  T.~He, Z.~Zhang, B.~Sch{\"o}lkopf, T.~Brox, and F.~Locatello.
\newblock Bridging the gap to real-world object-centric learning.
\newblock In \emph{The Eleventh International Conference on Learning
  Representations}, 2023.

\bibitem[Shah et~al.(2019)Shah, Chen, Rohrbach, and Parikh]{shah2019cycle}
M.~Shah, X.~Chen, M.~Rohrbach, and D.~Parikh.
\newblock Cycle-consistency for robust visual question answering.
\newblock \emph{CoRR}, abs/1902.05660, 2019.
\newblock URL \url{http://arxiv.org/abs/1902.05660}.

\bibitem[Singh et~al.(2022)Singh, Deng, and Ahn]{singh2022illiterate}
G.~Singh, F.~Deng, and S.~Ahn.
\newblock Illiterate {DALL}-e learns to compose.
\newblock In \emph{International Conference on Learning Representations}, 2022.

\bibitem[van~den Oord et~al.(2017)van~den Oord, Vinyals, and
  kavukcuoglu]{oord2017neural}
A.~van~den Oord, O.~Vinyals, and k.~kavukcuoglu.
\newblock Neural discrete representation learning.
\newblock In I.~Guyon, U.~V. Luxburg, S.~Bengio, H.~Wallach, R.~Fergus,
  S.~Vishwanathan, and R.~Garnett, editors, \emph{Advances in Neural
  Information Processing Systems}, volume~30. Curran Associates, Inc., 2017.
\newblock URL
  \url{https://proceedings.neurips.cc/paper/2017/file/7a98af17e63a0ac09ce2e96d03992fbc-Paper.pdf}.

\bibitem[Wah et~al.(2011)Wah, Branson, Welinder, Perona, and Belongie]{birds}
C.~Wah, S.~Branson, P.~Welinder, P.~Perona, and S.~Belongie.
\newblock Caltech-ucsd birds 200.
\newblock Technical Report CNS-TR-2011-001, California Institute of Technology,
  2011.

\bibitem[Wang et~al.(2013)Wang, Huang, and Guibas]{wang2013coseg}
F.~Wang, Q.~Huang, and L.~J. Guibas.
\newblock Image co-segmentation via consistent functional maps.
\newblock In \emph{2013 IEEE International Conference on Computer Vision},
  pages 849--856, 2013.
\newblock \doi{10.1109/ICCV.2013.110}.

\bibitem[Wang et~al.(2014)Wang, Huang, Ovsjanikov, and Guibas]{wandg2014coseg}
F.~Wang, Q.~Huang, M.~Ovsjanikov, and L.~J. Guibas.
\newblock Unsupervised multi-class joint image segmentation.
\newblock In \emph{2014 IEEE Conference on Computer Vision and Pattern
  Recognition}, pages 3142--3149, 2014.
\newblock \doi{10.1109/CVPR.2014.402}.

\bibitem[Wang et~al.(2022{\natexlab{a}})Wang, Liang, Shen, Gool, and
  Wang]{Wang2022CounterfactualCL}
H.~Wang, W.~Liang, J.~Shen, L.~V. Gool, and W.~Wang.
\newblock Counterfactual cycle-consistent learning for instruction following
  and generation in vision-language navigation.
\newblock \emph{2022 IEEE/CVF Conference on Computer Vision and Pattern
  Recognition (CVPR)}, pages 15450--15460, 2022{\natexlab{a}}.

\bibitem[Wang et~al.(2019)Wang, Song, Ma, Zhou, Liu, and
  Li]{wang2019unsupervised}
N.~Wang, Y.~Song, C.~Ma, W.~Zhou, W.~Liu, and H.~Li.
\newblock Unsupervised deep tracking.
\newblock \emph{CoRR}, abs/1904.01828, 2019.
\newblock URL \url{http://arxiv.org/abs/1904.01828}.

\bibitem[Wang et~al.(2022{\natexlab{b}})Wang, Shen, Hu, Yuan, Crowley, and
  Vaufreydaz]{Wang2022SelfSupervisedTF}
Y.~Wang, X.~Shen, S.~X. Hu, Y.~Yuan, J.~L. Crowley, and D.~Vaufreydaz.
\newblock Self-supervised transformers for unsupervised object discovery using
  normalized cut.
\newblock \emph{2022 IEEE/CVF Conference on Computer Vision and Pattern
  Recognition (CVPR)}, pages 14523--14533, 2022{\natexlab{b}}.

\bibitem[Wang et~al.(2023)Wang, Shou, and Zhang]{wang2023objectcentric}
Z.~Wang, M.~Z. Shou, and M.~Zhang.
\newblock Object-centric learning with cyclic walks between parts and whole,
  2023.

\bibitem[Watters et~al.(2019)Watters, Matthey, Burgess, and
  Lerchner]{watters2019spatial}
N.~Watters, L.~Matthey, C.~P. Burgess, and A.~Lerchner.
\newblock Spatial broadcast decoder: {A} simple architecture for learning
  disentangled representations in vaes.
\newblock \emph{CoRR}, abs/1901.07017, 2019.
\newblock URL \url{http://arxiv.org/abs/1901.07017}.

\bibitem[Wilson and Snavely(2013)]{kule2013network}
K.~Wilson and N.~Snavely.
\newblock Network principles for sfm: Disambiguating repeated structures with
  local context.
\newblock In \emph{2013 IEEE International Conference on Computer Vision},
  pages 513--520, 2013.
\newblock \doi{10.1109/ICCV.2013.69}.

\bibitem[Yoon et~al.(2023)Yoon, Wu, Bae, and Ahn]{yoon2023investigation}
J.~Yoon, Y.-F. Wu, H.~Bae, and S.~Ahn.
\newblock An investigation into pre-training object-centric representations for
  reinforcement learning, 2023.

\bibitem[Zach et~al.(2010)Zach, Klopschitz, and
  Pollefeys]{zach2010disambiguating}
C.~Zach, M.~Klopschitz, and M.~Pollefeys.
\newblock Disambiguating visual relations using loop constraints.
\newblock In \emph{2010 IEEE Computer Society Conference on Computer Vision and
  Pattern Recognition}, pages 1426--1433, 2010.
\newblock \doi{10.1109/CVPR.2010.5539801}.

\bibitem[Zhou et~al.(2015{\natexlab{a}})Zhou, Lee, Yu, and
  Efros]{zhou2015flowweb}
T.~Zhou, Y.~J. Lee, S.~X. Yu, and A.~A. Efros.
\newblock Flowweb: Joint image set alignment by weaving consistent, pixel-wise
  correspondences.
\newblock In \emph{2015 IEEE Conference on Computer Vision and Pattern
  Recognition (CVPR)}, pages 1191--1200, 2015{\natexlab{a}}.
\newblock \doi{10.1109/CVPR.2015.7298723}.

\bibitem[Zhou et~al.(2016)Zhou, Kr{\"{a}}henb{\"{u}}hl, Aubry, Huang, and
  Efros]{zhou2016learning}
T.~Zhou, P.~Kr{\"{a}}henb{\"{u}}hl, M.~Aubry, Q.~Huang, and A.~A. Efros.
\newblock Learning dense correspondence via 3d-guided cycle consistency.
\newblock \emph{CoRR}, abs/1604.05383, 2016.
\newblock URL \url{http://arxiv.org/abs/1604.05383}.

\bibitem[Zhou et~al.(2015{\natexlab{b}})Zhou, Zhu, and
  Daniilidis]{zhou2015multi}
X.~Zhou, M.~Zhu, and K.~Daniilidis.
\newblock Multi-image matching via fast alternating minimization.
\newblock In \emph{2015 IEEE International Conference on Computer Vision
  (ICCV)}, pages 4032--4040, 2015{\natexlab{b}}.
\newblock \doi{10.1109/ICCV.2015.459}.

\bibitem[Zhu et~al.(2017)Zhu, Park, Isola, and Efros]{zhu2017unpaired}
J.~Zhu, T.~Park, P.~Isola, and A.~A. Efros.
\newblock Unpaired image-to-image translation using cycle-consistent
  adversarial networks.
\newblock \emph{CoRR}, abs/1703.10593, 2017.
\newblock URL \url{http://arxiv.org/abs/1703.10593}.

\bibitem[Zoran et~al.(2021)Zoran, Kabra, Lerchner, and Rezende]{zorna2021parts}
D.~Zoran, R.~Kabra, A.~Lerchner, and D.~J. Rezende.
\newblock Parts: Unsupervised segmentation with slots, attention and
  independence maximization.
\newblock In \emph{2021 IEEE/CVF International Conference on Computer Vision
  (ICCV)}, pages 10419--10427, 2021.
\newblock \doi{10.1109/ICCV48922.2021.01027}.

\end{thebibliography}
